\documentclass[preprint,authoryear]{elsarticle}

\usepackage{lineno,hyperref}
\modulolinenumbers[5]

\usepackage{wrapfig}
\usepackage{amsmath,amsbsy,amsfonts,amssymb,amsthm,dsfont,units}
\usepackage{xspace}
\usepackage{graphicx}
\usepackage{algorithm,algorithmic,mathtools}
\usepackage{color,cases}
\usepackage{tikz}
\usetikzlibrary{calc,shapes}
\usepackage{subfigure}
\usepackage{multirow}
\usepackage{makecell}
\usepackage{rotating}

\usepackage{hyperref}
\hypersetup{
colorlinks = true,
citecolor = {blue},
citebordercolor = {white}
}

\usepackage[T1]{fontenc}
\usepackage[utf8x]{inputenc}
\usepackage{units}
\usepackage{amsmath}
\usepackage{amsthm}
\usepackage{graphicx}
\usepackage{esint}
\usepackage{bm}
\usepackage{algorithm}
\usepackage{defs_151115}
\usepackage{algorithmic}
\usepackage{amsthm}
\usepackage{xargs}[2008/03/08]
\usepackage{amsfonts}       
\usepackage{microtype}      

\newcommand{\zz}{{\bf z}}
\newcommand{\slerp}{\mathrm{slerp}}
\newcommand{\ZZ}{\mathcal{Z}}

\usepackage{relsize}

\usepackage{pifont}
\newcommand{\etal}{\mbox{\emph{et al.\ }}}

\usepackage{mathtools}

\usepackage{url}
\usepackage{amsfonts}
\usepackage{nicefrac}
\usepackage{microtype}

\usepackage{graphicx}

\usepackage{colortbl} 
\definecolor{header_color}{rgb}{0.74,0.88,0.91}
\definecolor{even_color}{rgb}{0.9,0.9,0.9}
\definecolor{subheader_color}{rgb}{0.85,0.93,0.95}
\definecolor{childheader_color}{rgb}{1.0,0.93,0.87}

\definecolor{ccolor_best}{rgb}{1.0,0.9,0.9}
\definecolor{ccolor_wrong}{rgb}{1.0,0.85,0.85}

\usepackage{caption}

\usepackage{thmtools}

\declaretheoremstyle[%
  spaceabove=-6pt,%
  spacebelow=6pt,%
  headfont=\normalfont\itshape,%
  postheadspace=1em,%
  qed=\qedsymbol%
]{mystyle}

\global\long\def\vt#1{\mathbf{#1}}

\global\long\def\x{\mathrm{x}}

\global\long\def\by{\boldsymbol{y}}

\global\long\def\bw{\vt w}

\global\long\def\and{\cap}

\global\long\def\ess{\mathbb{E}}

\newcommandx\ESS[2][usedefault, addprefix=\global, 1=]{\underset{#2}{\ess}\left[#1\right]}

\vspace{-2mm}

\def\ie{\emph{i.e.~}}
\def\eg{\emph{e.g.~}}
\def\etal{{\em et al.}}
\def\aka{{\em a.k.a~}}

\begin{document}

\begin{frontmatter}

\title{Pros and Cons of GAN Evaluation Measures:\\ New Developments}

\author{Ali Borji \\ aliborji@gmail.com } 




\begin{abstract}
This work is an update of a previous paper on the same topic published a few years ago~\citep{borji2019pros}. With the dramatic progress in generative modeling, a suite of new quantitative and qualitative techniques to evaluate models has emerged. Although some measures such as Inception Score, Fr\'echet Inception Distance, Precision-Recall, and Perceptual Path Length are relatively more popular, GAN evaluation is not a settled issue and there is still room for improvement. Here, I describe new dimensions that are becoming important in assessing models (\eg bias and fairness)
and discuss the connection between GAN evaluation and deepfakes. These are important areas of concern in the machine learning community today and progress in GAN evaluation can help mitigate them.  
\end{abstract}


\end{frontmatter}




\section{Introduction}



Generative models have revolutionized AI and are now being widely adopted\footnote{For a list of GAN applications, please refer to \href{https://machinelearningmastery.com/impressive-applications-of-generative-adversarial-networks/}{here}.}. They are remarkably effective at synthesizing strikingly realistic and diverse images, and learn to do so in an unsupervised manner~\citep{wang2019generative,liu2021generative,bond2021deep}. Models such as generative adversarial networks (GANs)~\citep{goodfellow2014generative}, autoregressive models such as PixelCNNs~\citep{oord2016conditional}, variational autoencoders (VAEs)~\citep{kingma2013auto}, and recently transformer GANs~\citep{hudson2021generative,jiang2021transgan} have been constantly improving the state of the art in image generation. See for example~\cite{brock2018large,karras2019style,ramesh2021zero} and~\cite{razavi2019generating}. 

A key difficulty in generative modeling is evaluating performance, \ie how good is a model in approximating a data distribution? Quantitative evaluation\footnote{Generative models are commonly evaluated in terms of fidelity (how realistic a generated image is) and diversity (how well generated samples capture the variations in real data) of the learned distribution. Generative models, however, can also tested against other criteria such as linear separability, mode collapse, etc. For a complete list of desired properties of GANs, please see~\cite{borji2019pros}.} of generative models of images and video is an open problem. Here, I give a summary and discussion of the latest progress, since~\cite{borji2019pros}, in this field covering newly proposed measures, benchmarks, and GAN visualization and error diagnosis techniques.

\subsection{Background}
Evaluation of generative models is an active area of research. Several quantitative and qualitative measures have been proposed so far. In this section, I review some of the influential measures that have inspired the researchers.



A classic approach is to compare the log-likelihoods of models. This approach, however, has several shortcomings. A model can achieve high likelihood, but low image quality, and conversely, low likelihood and high image quality. Also, kernel density estimation in high-dimensional spaces is very challenging~\citep{theis2015note}.




The two most common GAN evaluation measures are Inception Score (IS) and Fr\'echet Inception Distance (FID). They rely on a pre-existing classifier (InceptionNet) trained on ImageNet. IS~\citep{salimans2016improved} computes the KL divergence between the conditional class distribution and the marginal class distribution over the generated data. FID~\citep{heusel2017gans} calculates the Wasserstein-2 (\aka Fr\'echet) distance between multivariate Gaussians fitted to the embedding space of the Inception-v3 network of generated and real images\footnote{To learn how to implement FID, please refer to \href{https://machinelearningmastery.com/how-to-implement-the-Fr\'echet-inception-distance-fid-from-scratch/}{here}.}. 

IS does not capture intra-class diversity, is insensitive to the prior distribution over labels (hence is biased towards ImageNet dataset and Inception model\footnote{See also~\href{https://medium.com/octavian-ai/a-simple-explanation-of-the-inception-score-372dff6a8c7a}{here}.}), and is very sensitive to model parameters and implementations~\citep{barratt2018note,odena2019open}. It requires a large sample size to be reliable. A simple class-conditional model that memorizes one example per ImageNet class achieves a high IS. Using a 1D example,~\cite{barratt2018note} show that the true underlying data distribution may achieve a lower IS than other distributions.


FID has been widely adopted because of its consistency with human inspection and sensitivity to small changes in the real distribution (\eg slight blurring or small artifacts in synthesized images). FID, unlike IS, can detect intra-class mode collapse\footnote{Mode collapse happens when a generator produces data from only a few modes of the target distribution, thus failing to generate diverse enough outputs. Please see \href{https://wandb.ai/authors/DCGAN-ndb-test/reports/Measuring-Mode-Collapse-in-GANs--VmlldzoxNzg5MDk}{here}.}. The Gaussian assumption in FID calculation, however, might not hold in practice. A major drawback with FID is its high bias. The sample size to calculate FID has to be large enough (usually above 50K). Smaller sample sizes can lead to over-estimation of the actual FID~\citep{chong2020effectively}. 

While single-value metrics like IS and FID successfully capture many aspects of the generator, they lump quality and diversity evaluation and are therefore not ideal for diagnostic purposes (Precision-Recall addresses this shortcoming). Further, they continue to have a blind spot for image quality~\citep{karras2019style}.

Another popular approach to evaluate GANs is manual inspection (\eg~\cite{denton2015deep,zhou2019hype}). While this approach can directly tell us about the image quality, it is limited in assessing sample diversity. It is also subjective as the reviewers may incorporate their own biases and opinions. It also requires knowledge of what is realistic and what is not for the target domain which might be hard to learn in some domains (\eg medical domain\footnote{For example, to tell whether a GAN can generate X-rays with pneumonia in them, it is better to ask doctors rather than Amazon Mechanical Turk workers!}). Further, it is constrained by the number of images that can be reviewed in a reasonable time prohibiting its usage during model development. Finally, due to the subtle differences in experimental protocols (\eg user interface design, fees, duration), it is often difficult to replicate the results across different publications.


\subsection{Organization of the Paper}
This paper is organized as follows. Recently proposed evaluation measures are categorized under quantitative and qualitative measures and are covered in sections 2 and 3, respectively\footnote{Some methods may fall in both categories.}. In section 4, I then discuss some benchmark and model comparison studies as well as analysis works. Finally, section 5 highlights some major points and offers suggestions for future research in GAN evaluation. Note that some of the works reviewed here have not been accepted in prior published work and may have not been peer reviewed. Therefore, they should be treated with additional caution in the absence of widespread usage.

\section{New Quantitative GAN Evaluation Measures}

\subsection{Specialized Variants of Fr\'echet Inception Distance and Inception Score}

\subsubsection{Spatial FID (sFID)}
\cite{nash2021generating} propose a variant of FID called sFID that uses spatial features rather than the standard pooled features. They compute FID using both the standard pool3 inception features and the first 7 channels from the intermediate mixed 6/conv feature maps. 
The reason to include pool3 is because it compresses spatial information to a large extent, making it less sensitive to spatial variability. The reason to include the intermediate mixed 6/conv is because it provides a sense of spatial distributional similarity between models. 

\subsubsection{Class-aware FID (CAFD) and Conditional FID}
\cite{liu2018improved} argue against the
single-manifold Gaussian assumption in FID, and employ a Gaussian mixture model (GMM) to better fit the feature distribution and to include class information. They compute Fr\'echet distance in each of the $K$ classes and average the results to obtain CAFD:
\[  
CAFD(P_r, P_g)=\frac{1}{K}\sum_{i=1}^K{||\mu^r_i-\mu^g_i||+Tr(C^r_i+C^g_i-2(C^r_iC^g_i)^\frac{1}{2})},
\]
where $\mu_i$ and $C_i$ are the mean and the covariance matrix of class $i$, respectively.

\cite{soloveitchik2021conditional} introduce a conditional variant of FID for evaluating conditional generative models. Instead of having two distributions, they consider two classes of distributions conditioned on a common input. 

\subsubsection{Fast FID}



\cite{mathiasen2020fast} propose a method to speed up the FID computation in order to make it computationally tractable such that it can be used as a loss function for training GANs. The real data samples do not change during training, so their Inception encodings need to be computed once. Therefore, reducing the number of fake samples lowers the time to compute the Inception encodings. The bottleneck, however, is computing $\tr(\sqrt{\Sigma_1\Sigma_2})$ in the FID formula. $\Sigma_1$ and $\Sigma_2$ are the covariance matrices of Gaussians fitted to the real and the generated data. FastFID circumvents this issue without explicitly computing $\sqrt{\Sigma_1\Sigma_2}$. Depending on the number of fake samples, FastFID can speed up FID computation 25 to 500 times. Notice that FastFID makes backpropagating over FID faster, and as such it is not a metric, rather an optimization method.


\subsubsection{Memorization-informed FID (MiFID)}
\cite{bai2021training} 
conduct the first generative model competition\footnote{\url{https://www.kaggle.com/c/generative-dog-images}}, where participants were invited to generate realistic dog images given 20,579 images of dogs from ImageNet. They modified the FID score to penalizes models producing images too similar to the
training set as
\[
\text{MiFID}(S_g, S_t) = m_\tau(S_g, S_t) \cdot \text{FID}(S_g, S_t)
\]
where $S_g$ is the generated set and $S_t$ is the original training set. $m_\tau$ is the memorization penalty which is based on thresholding the memorization distance $s$ of 
generated and true distribution defined as:
\[
    s(S_g, S_t) = \frac{1}{|S_g|} \sum_{x_g \in S_g} \min_{x_t \in S_t} 
    \bigg{(} 1 -  \frac{|\langle x_g, x_t \rangle|}{|x_g| \cdot |x_t|} \bigg{)} \]
\[    
    m_\tau(S_g, S_t) = \left\{\begin{array}{ll} 
    {\frac{1}{s(S_g, S_t) + \epsilon} \quad (\epsilon \ll 1),} & {\text { if } s(S_g, S_t) < \tau} \\
    {1,} & {\text { otherwise }}
    \end{array}\right.
    \]

Intuitively, lower memorization distance is associated with more severe training sample memorization.

\subsubsection{Unbiased FID and IS}
\cite{chong2020effectively} show that FID and IS are biased in the sense that their expected value, computed using a finite sample set, is not their true value. They find that bias depends on the number of images used for calculating the score and the generators themselves, making objective comparisons between models difficult. To mitigate the issue, they propose an extrapolation approach to obtain a bias-free estimate of the scores, called $\overline{FID}_\infty$ and $\overline{IS}_\infty$, computed with an infinite number of samples. 
These extrapolated scores are simple, drop-in replacements for the finite sample scores\footnote{\url{https://github.com/mchong6/FID_IS_infinity}}.

\subsubsection{Clean FID}
\cite{parmar2021buggy} examine the sensitivity of FID score (and also other scores such as KID) to inconsistent and often incorrect implementations across different image processing libraries when training and evaluating generative models. FID score is widely used to evaluate generative models, but each FID implementation uses a different low-level image processing method. Image resizing functions in commonly-used deep learning libraries often introduce aliasing artifacts. They observe that numerous subtle choices need to be made for FID calculation and a lack of consistencies in these choices can lead to vastly different FID scores. To address these challenges, they introduce a standardized protocol, and provide an easy-to-use FID evaluation library called cleanfid\footnote{\url{github.com/GaParmar/clean-fid}}.



\subsubsection{Fr\'echet Video Distance (FVD)}
\cite{unterthiner2019fvd} propose an extension of the FID, called FVD, to evaluate generative models of video.
In addition to the quality of each frame, FVD captures the temporal coherence in a video. To obtain a suitable feature video representation, they used a pre-trained network (the Inflated 3D Convnet~\citep{carreira2017quo} pre-trained on Kinetics-400 and Kinetics-600 datasets) that considers the temporal coherence of the visual content across a sequence of frames. The I3D network generalizes the Inception architecture to sequential data, and is trained to perform action-recognition on the Kinetics dataset consisting of human-centered YouTube videos. FVD considers a distribution over videos, thereby avoiding the drawbacks of frame-level measures. 
  
Some other measures for video synthesis evaluation include Average Content Distance~\citep{tulyakov2018mocogan} (the average L2 distance among all consecutive frames in a video), Cumulative Probability Blur Detection (CPBD)~\citep{narvekar2009no}, Frequency Domain Blurriness Measure (FDBM)~\citep{de2013image}, Pose Error~\citep{yang2018pose}, and Landmark Distance (LMD)~\citep{chen2018lip}. For more details on video generation and evaluation, please consult~\cite{oprea2020review}.

In addition to above, Fr\'echet Audio Distance (FAD)~\citep{roblek2019fr} and Fr\'echet ChemNet Distance (FCD)~\citep{preuer2018frechet}
have been suggested for evaluation of music enhancement and molecule generation algorithms.

\subsection{Methods based on Self-supervised Learned Representations}

FID scores computed with classification-pretrained embeddings have been shown to correlate well with human evaluations~\citep{heusel2017gans}. Meanwhile, they can be misleading as they are biased towards ImageNet. 
On non-ImageNet datasets, FID can result in inadequate evaluation. \cite{morozov2020self} advocate for using self-supervised representations to evaluate GANs on the established non-ImageNet benchmarks\footnote{\url{https://github.com/stanis-morozov/self-supervised-gan-eval}}. They show that representations, typically obtained via contrastive or clustering-based approaches, provide better transfer to new tasks and domains, and thus can serve as more universal embeddings of natural images. Further, they demonstrate that self-supervised representations produce a more reasonable ranking of models and often improve the sample efficiency of FID.





\subsection{Methods based on Analysing Data Manifold}
These methods measure disentanglement in the representations learned by generative models, and are useful for improving the generalization, robustness, and interpretability of models.

\subsubsection{Local Intrinsic Dimensionality (LID)} 
\begin{figure}[t]
\begin{center}
\includegraphics[width=1\linewidth]{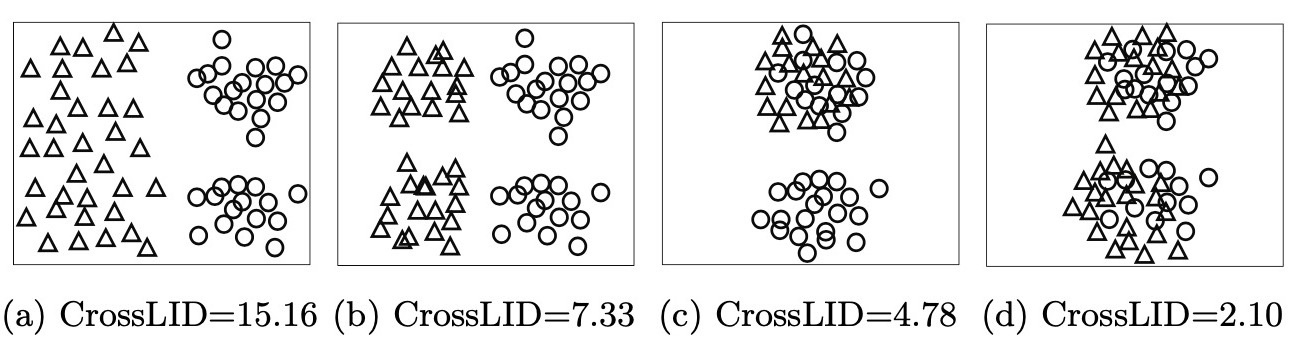}
\caption{Four 2D examples illustrating how generated samples by a GAN (triangles)
relate to real data samples from a bimodal Gaussian (circles), together with
CrossLID scores. (a) generated data distributed uniformly, spatially far from the
real data. (b) generated data with two modes, spatially far from the real data (c) generated data associated with only one mode of the real data. (d) generated data associated with both modes of the real data (the desired situation). The lower the CrossLID, the better. Figure compiled from~\cite{barua2019quality}.}
\label{fig:barua2019quality}
\end{center}
\end{figure}

\cite{barua2019quality} introduce a new evaluation measure called ``CrossLID'' to assess the local intrinsic dimensionality (LID) of real-world data with respect to neighborhoods found in GAN-generated samples. Intuitively, CrossLID measures the degree to which the manifolds of two data distributions coincide with each other. The idea behind CrossLID is depicted in Fig.~\ref{fig:barua2019quality}. Barua \etal~compare CrossLID with other measures and show that it a) is strongly correlated with the progress of GAN training, b) is sensitive to mode collapse, c) is robust to small-scale noise and image transformations, and 4) is robust to sample size. It is not clear whether this measure can be applied to complex and high dimensional data where it is hard to define local dimensionality.


\subsubsection{Intrinsic Multi-scale Distance (IMD)}
\begin{figure}[t]
\begin{center}
  \includegraphics[width=.3\linewidth]{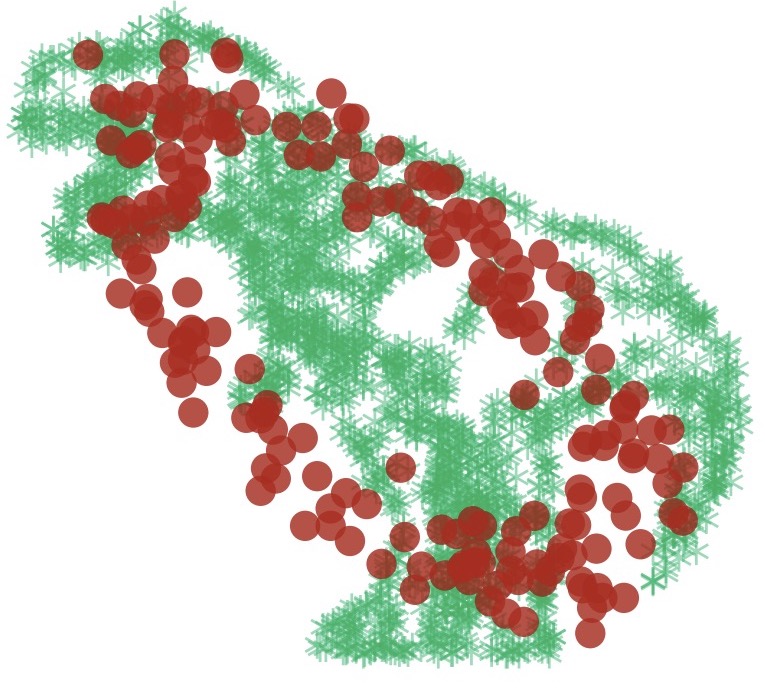}
  \caption{The motivation behind the Intrinsic Multi-scale Distance (IMD) score. Here, the two distributions have the same first 3 moments. Since FID and KID are insensitive to higher moments, they can not distinguish the two distributions, whereas the IMD score can. Figure from~\cite{tsitsulin2019shape}.}
\label{fig:tsitsulin2019shape}   
\end{center}
\end{figure}

\cite{tsitsulin2019shape} argue that current evaluation measures only reflect the first two (mean and covariance) or three moments of distributions. 
As illustrated in Fig.~\ref{fig:tsitsulin2019shape}, FID and KID (Kernel Inception Distance~\citep{binkowski2018demystifying})\footnote{KID is refereed to as MMD in~\cite{borji2019pros}.} are insensitive to the global structure of the data distribution. 
They propose a measure, called IMD, to take all data moments into account, and claim that IMD is intrinsic\footnote{It is invariant to isometric transformations of the manifold such as translation or rotation.} and multi-scale\footnote{It captures both local and global information.}. Tsitsulin \etal~experimentally show that their method is effective in discerning the structure of data manifolds even on unaligned data. IMD compares data distributions based on their geometry and is somewhat similar to the Geometry Score~\citep{khrulkov2018geometry}. 

\cite{barannikov2021manifold} introduce
Cross-Barcode(P,Q), given a pair of distributions in a high-dimensional space, tracks multiscale topology discrepancies between manifolds on which the distributions are concentrated. Based on the Cross-Barcode, they introduce the Manifold Topology Divergence score (MTop-Divergence) and apply it to assess the performance of deep generative models in various domains: images, 3D-shapes, and time-series. Their proposed method is domain agnostic and does not rely on pre-trained networks


\subsubsection{Perceptual Path Length (PPL)} 
\begin{figure}[t]
\begin{center}
   \includegraphics[width=.7\linewidth]{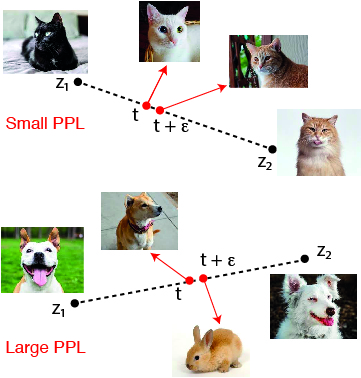}
   \caption{Illustration of PPL. Small perturbations around a point result in smaller average perceptual distance for a model that has learned a disentangled representation.}
\label{fig:ppl2}   
\end{center}
\end{figure}

\begin{figure}[t]
\begin{center}
   \includegraphics[width=.9\linewidth]{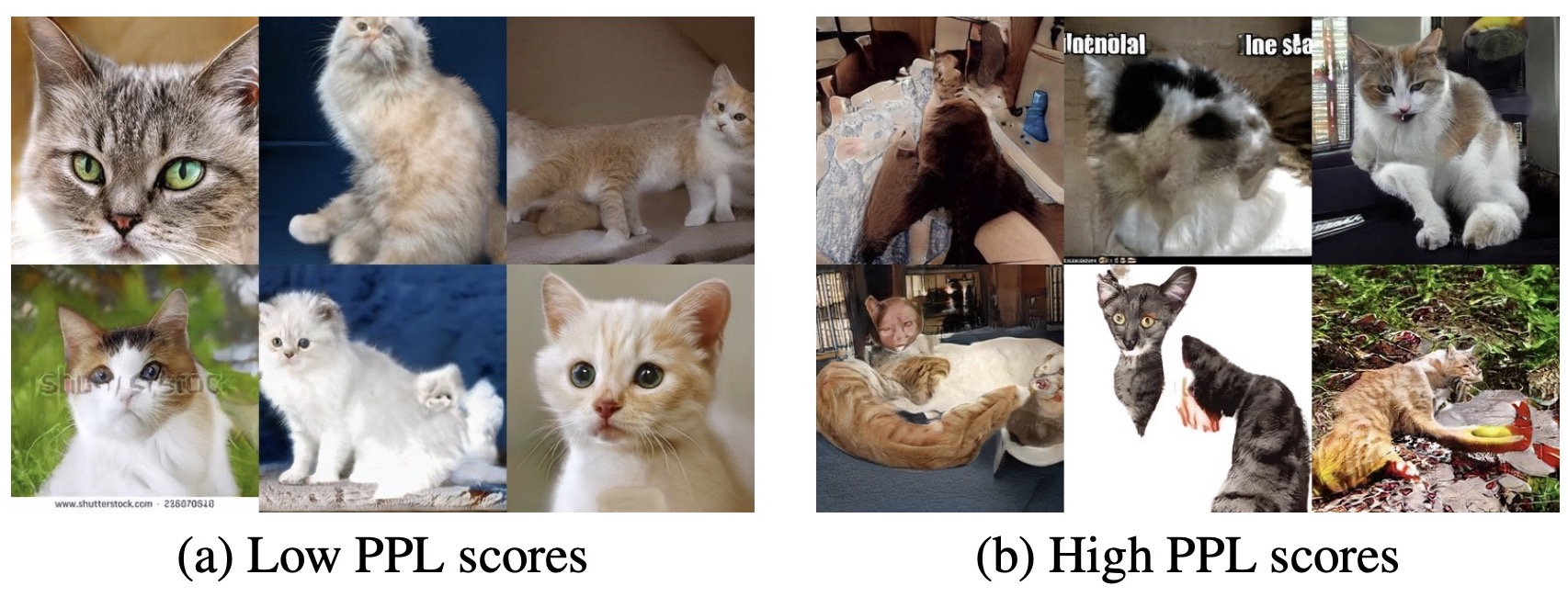}
   \caption{Correlation between PPL and image quality (generated by StyleGAN). Panels (a) and (b) show random examples with low and high (per-image) PPL. PPL score is capable of capturing the consistency of the images. The lower the PPL, the better. Figure from~\cite{karras2020analyzing}.}
\label{fig:ppl}   
\end{center}
\end{figure}

\begin{figure}[htbp]
\begin{center}
\includegraphics[width=.9\linewidth]{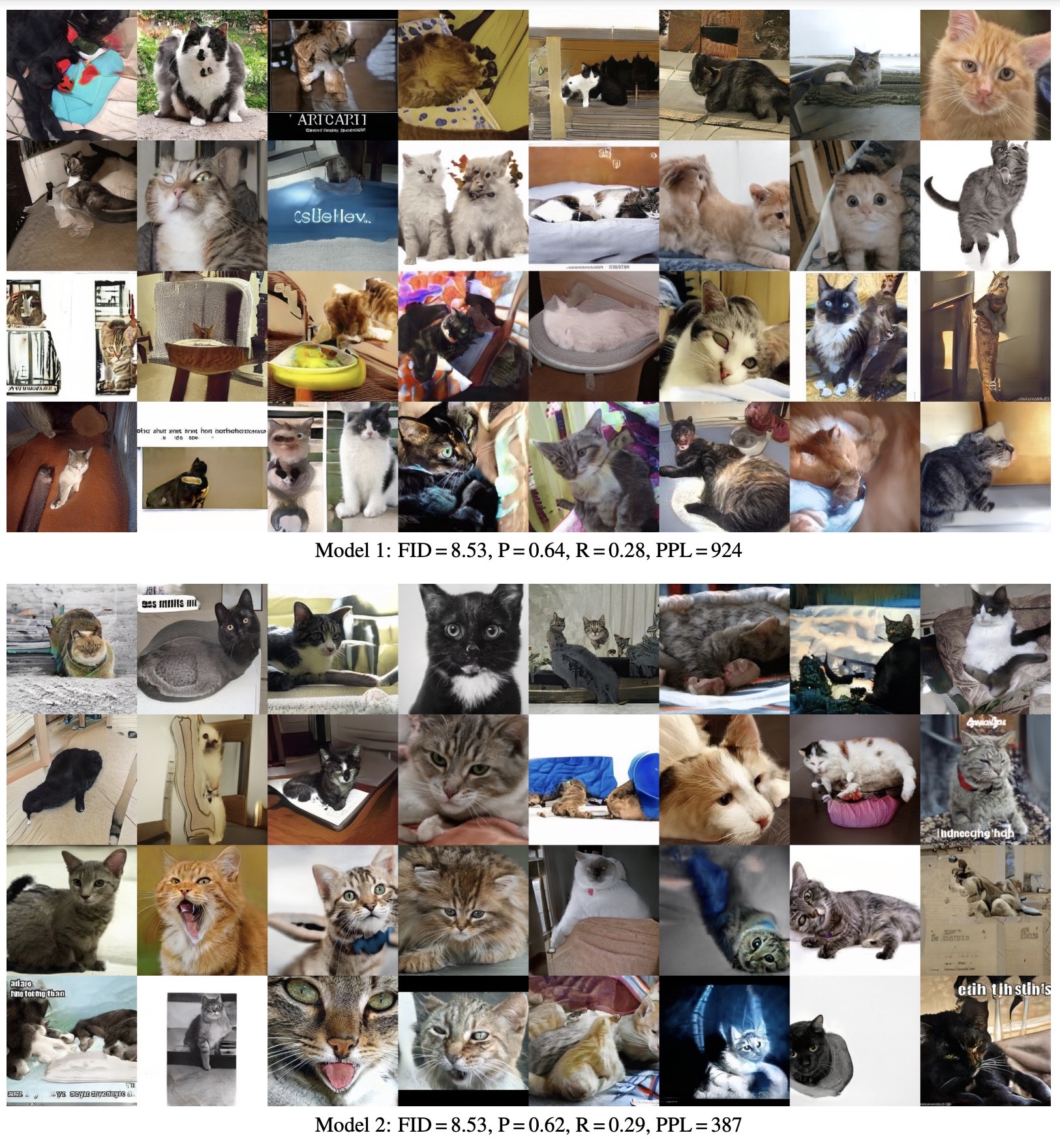}
\caption{Synthesized samples from two generative models trained on LSUN~\citep{yu2015lsun} CAT without truncation (\ie drawing latent vectors from a truncated or shrunk sampling space to improve average image quality, at the expense of loosing diversity~\citep{kingma2013auto,brock2018large,karras2020analyzing}). FID, precision (P), and recall (R) are similar for the two models, even though the latter produces cat-shaped objects more often. Perceptual path length (PPL) shows a clear preference for model 2. 
Figure from~\cite{karras2020analyzing}.}
\label{fig:karras2020analyzingCAT}
\end{center}
\end{figure}

First introduced in StyleGAN3~\citep{karras2019style}, 
PPL measures whether and how much the latent space of a generator is entangled (or if it is smooth and the factors of variation are properly separated\footnote{For example, features that are absent in either endpoint may appear in the middle of a linear interpolation path between two random inputs.}). Intuitively, a less curved latent space should result in perceptually smoother transition than a highly curved latent space (See Fig.~\ref{fig:ppl2}). 
Formally, PPL is the empirical mean of the perceptual difference between consecutive images in the latent space $\ZZ$, over all possible endpoints:
\[
l_{\ZZ} = \mathbb{E}[{\displaystyle\frac{1}{\epsilon^2}}d\big(G(\slerp(\zz_1,\zz_2;\,t)), G(\slerp(\zz_1,\zz_2;\,t+\epsilon))\big)\Big]\textrm{,}
\]
where \mbox{$\zz_1,\zz_2\sim P(\zz), t\sim U(0,1)$}, $G$ is the generator, and $d(\cdot,\cdot)$ is the perceptual distance between the resulting images. $\slerp$ denotes the spherical interpolation~\citep{shoemake1985animating}, and $\epsilon=10^{-4}$ is the step size. The LPIPS (learned perceptual image patch similarity)~\citep{zhang2018unreasonable} can be used to measure the perceptual distance between two images. LPIPS is a weighted L2 difference between two VGG16~\citep{simonyan2014very} embeddings, where the weights are learned to make the metric agree with human perceptual similarity judgments. 
PPL captures semantics and image quality as shown in Fig.~\ref{fig:ppl}. 
Fig.~\ref{fig:karras2020analyzingCAT} illustrates the superiority of PPL over FID and Precision-Recall scores in comparing two models.


\subsubsection{Linear Separability in Latent Space}




\cite{karras2019style} propose another measure in StyleGAN3 to quantify the latent space disentanglement by measuring how well the latent-space points can be separated into two distinct sets via a linear hyperplane. They argue that if a latent space is sufficiently disentangled, it should be possible to find directions that consistently correspond to individual factors of variation. 

First, some auxiliary classification networks for a number of binary attributes are trained (\eg to distinguish male and female faces). To measure the separability
of an attribute:
\begin{enumerate}
    \item Generate 200K images with $\zz\sim P(\zz)$ and classify them using an auxiliary classification network with label $Y$,
    \item Sort the samples according to classifier confidence and remove the least confident half, yielding 100K labeled latent-space vectors,
    \item Fit a linear SVM to predict the label $X$ based only on the latent-space points and classify the points by this plane,
    \item Compute the conditional entropy ${\mathrm H}(Y|X)$ where $X$ represents the classes predicted by the SVM and $Y$ represents the classes determined by the classifier. A low value suggests consistent latent space directions for the corresponding factor(s) of variation.
\end{enumerate}

The final separability score is $\exp(\sum_i{\mathrm H}(Y_i|X_i))$, where $i$ enumerates over a set of attributes (\eg $i=40$ over CelebA dataset). 





\subsection{Classification Accuracy Score (CAS) } 
\begin{figure*}[t]
\begin{center}
\includegraphics[width=1\linewidth]{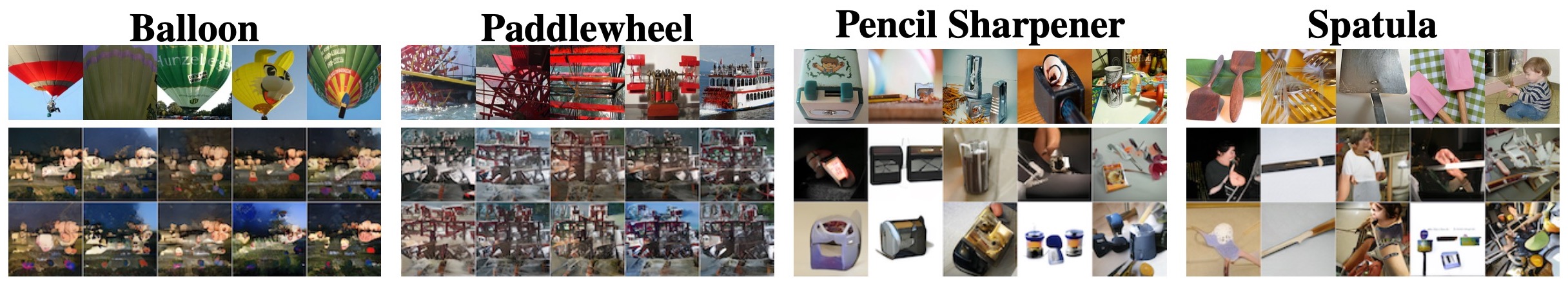}
\caption{CAS can identify classes for which a generative model, here BigGAN-deep~\citep{brock2018large}, fails to capture the data distribution (top row: real images, bottom two rows: generated samples). Figure compiled from~\cite{ravuri2019classification}.}
\label{fig:ravuri2019classification}
\end{center}
\end{figure*}

\cite{ravuri2019classification} propose a measure that is in essence similar to the ``Classification Performance'' discussed in~\cite{borji2019pros}.
They argue that if a generative model is learning the data distribution in a perceptually meaningful space then it should perform well in downstream tasks. They use class-conditional generative models from a number of generative models such as GANs and VAEs to infer the class labels of real data. They then train an image classifier using only synthetic data and use it to predict labels of real images in the test set. Their findings are as follows: a) when using a state-of-the-art GAN (BigGAN-deep by~\cite{brock2018large}), Top-1 and Top-5 accuracies decrease by 27.9\% and 41.6\%, respectively, compared to the original data, b) CAS automatically identifies particular classes for which generative models fail to capture the data distribution (Fig.~\ref{fig:ravuri2019classification}), and c) IS and FID are neither predictive of CAS, nor useful when evaluating non-GAN models.

\subsection{Non-Parametric Tests to Detect Data-Copying}
\begin{figure}[t]
\begin{center}
   \includegraphics[width=1\linewidth]{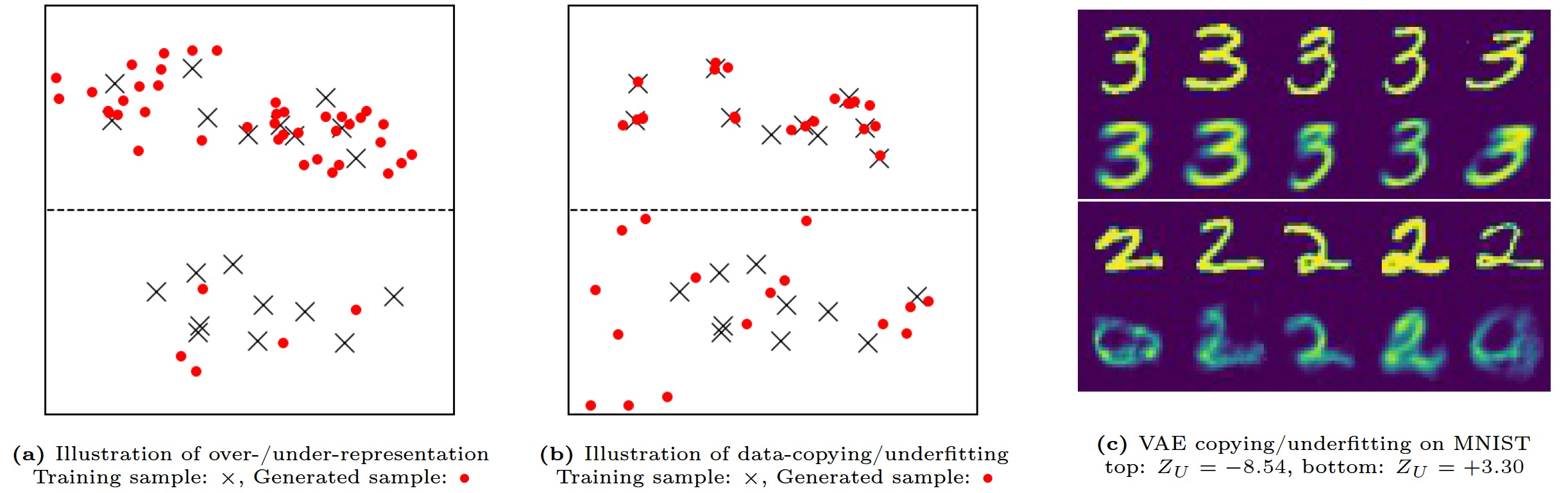}
   \caption{Illustration of the data-copying concept introduced in~\cite{meehan2020non}. Each panel depicts a single instance space partitioned into two regions. Panel (a) shows an over-represented region (top) and an under-represented region (bottom). This is the kind of overfitting evaluated by measures such as FID and Precision-Recall. Panel (b) shows a data-copied region (top) and an underfitted region (bottom). 
   Panel (c) shows VAE-generated and training samples from a data-copied (top) and underfitted (bottom) region over MNIST. In each 10-image strip, the bottom row provides random generated samples from the region and the top row shows their training nearest neighbors. Samples in the bottom region are on average further to their training nearest neighbor than held-out test samples in the region, and samples in the top region are closer, and thus ``data-copying''. 
Figure from~\cite{meehan2020non}.}
\label{fig:meehan2020non}
\end{center}
\end{figure}

\cite{meehan2020non} formalize a notion of overfitting called data-copying where a generative model memorizes the training samples or their small variations (Fig.~\ref{fig:meehan2020non}). They provide a three sample test for detecting data-copying that uses the training set, a separate held-out sample from the target distribution, and a generated sample from the model. The key insight is that an overfitted GAN generates samples that are on average closer to the training samples, therefore the average distance of generated samples to training samples is smaller than the corresponding distances between held-out test samples and the training samples. They also divide the instance space into cells and conduct their test separately in each cell. This is because, as they argue, generative models tend to behave differently in different regions of space.

\subsection{Measures that Probe Generalization in GANs}
\label{sec:generalization}
\begin{figure}[t]
\begin{center}
   \includegraphics[width=1\linewidth]{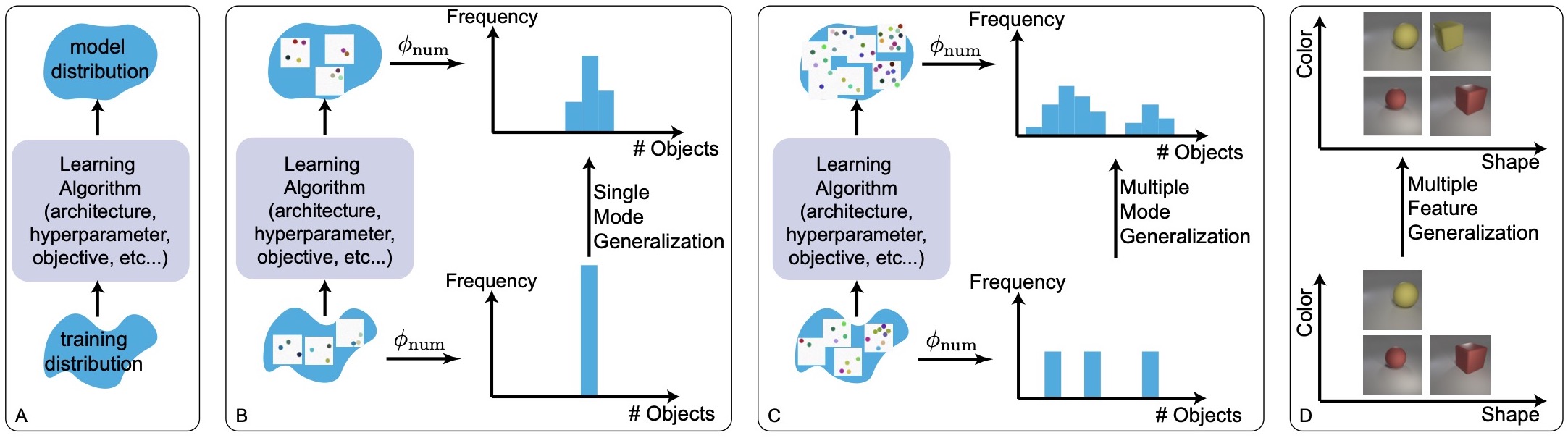}
   \caption{A) A generative model can be probed with carefully designed training data. Examining the learned distribution when training data B) takes a single value for a feature (\eg all training images have 3 objects), C) has multiple modes for a feature (\eg all training images have 2, 4 or 10 objects), or D) has multiple modes over multiple features. Figure compiled from~\cite{zhao2018bias}.}
   \label{fig:zhao2018bias}
\end{center}
\end{figure}

\begin{figure}[t]
\begin{center}
\includegraphics[width=1\linewidth]{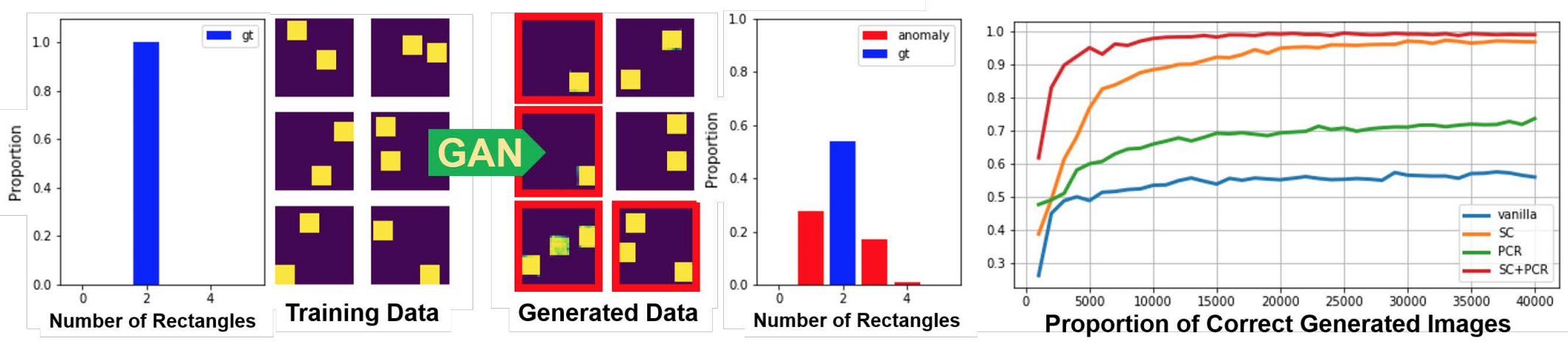}
\caption{\cite{xuan2019anomalous} show that training a GAN over images with exactly two rectangles results in a model that generates one, two, or three rectangles (anomalous ones shown in red). They also propose a model that generates a high fraction of correct images (the right-most panel).}
\label{fig:xuan2019anomalous}
\end{center}
\end{figure}

\cite{zhao2018bias}\footnote{Please see also the subsection on ``Evaluating Mode Drop and Mode Collapse'' in~\cite{borji2019pros}.} utilize carefully designed training datasets to characterize how existing models generate novel attributes and their combinations.
Some of their findings are as follows. When presented with a training set with all images having exactly 3 objects, both GANs and VAEs typically generate 2-5 objects
(Fig.~\ref{fig:zhao2018bias}.B). Over a multi-modal training distribution (\eg images with either 2 or 10 objects), the model acts as if it is trained separately on each mode, and it averages the two modes. When the modes are close to each other (\eg 2 and 4 objects), the learned distribution assigns higher probability to the mean number of objects (3 in this example), even though there was no image with 3 objects in the training set (Fig.~\ref{fig:zhao2018bias}.C). When the training set contains certain combinations (\eg red cubes but not yellow cubes; Fig.~\ref{fig:zhao2018bias}.D), the model memorizes the combinations in the training set when it contains a small number of them (\eg 20), and generates novel combinations when there is more variety (\eg 80). \cite{xuan2019anomalous} made a similar observation shown in Fig.~\ref{fig:xuan2019anomalous}. On a geometric-object training dataset where the number of objects for each training image is fixed, they found a variable number of objects in generated images. Some other studies that have investigated GAN generalization include~\cite{o2018evaluating,van2020investigating}.

\subsection{New Ideas based on Precision and Recall (P\&R)}
\begin{figure}[t]
\begin{center}
\includegraphics[width=\linewidth]{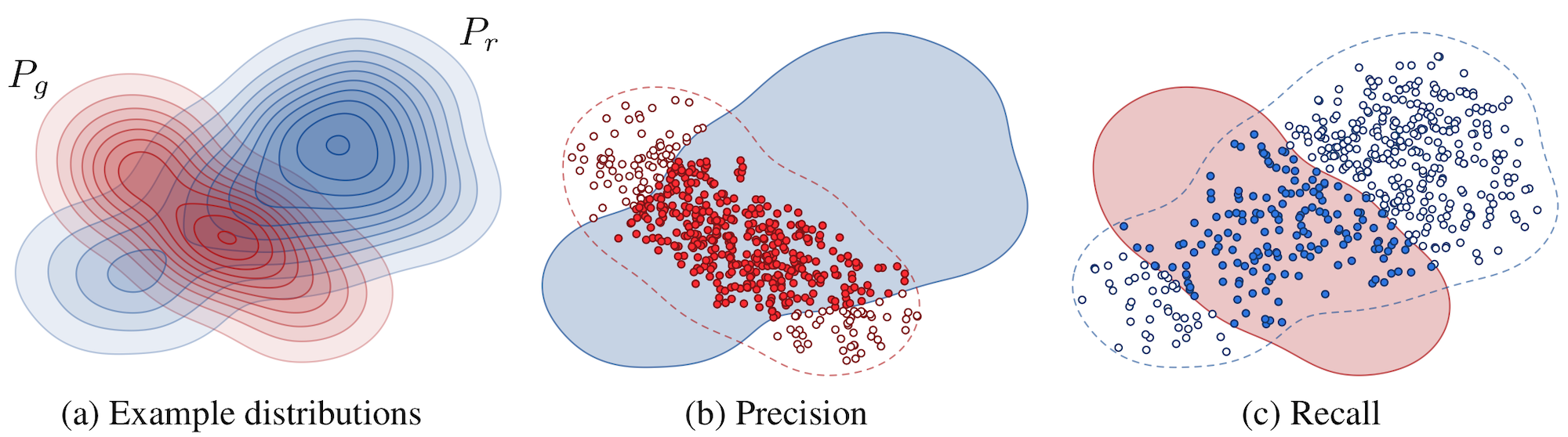}
\caption{(a) Illustration of precision-recall for distribution of real images $P_r$ (blue) and the distribution of generated images $P_g$ (red). (b) Precision is the probability that a random image from $P_g$ falls within the support of $P_r$. (c) Recall is the probability that a random image from $P_r$ falls within the support of $P_g$. Figure compiled from~\cite{kynkaanniemi2019improved}.} 
\label{fig:Tuomas}
\end{center}
\end{figure}

\cite{sajjadi2018assessing} propose to use precision and recall to explicitly quantify the trade off between quality (precision) and coverage (recall). Precision measures the similarity of generated instances to the real ones and recall measures the ability of a generator to synthesize all instances found in the training set (Fig.~\ref{fig:Tuomas}). P\&R curves can distinguish mode-collapse (poor recall) and bad quality (poor precision). Some new ideas based on P\&R are summarized below. 

\subsubsection{Density and Coverage} 
\begin{figure}[t]
\begin{center}
\includegraphics[width=1\linewidth]{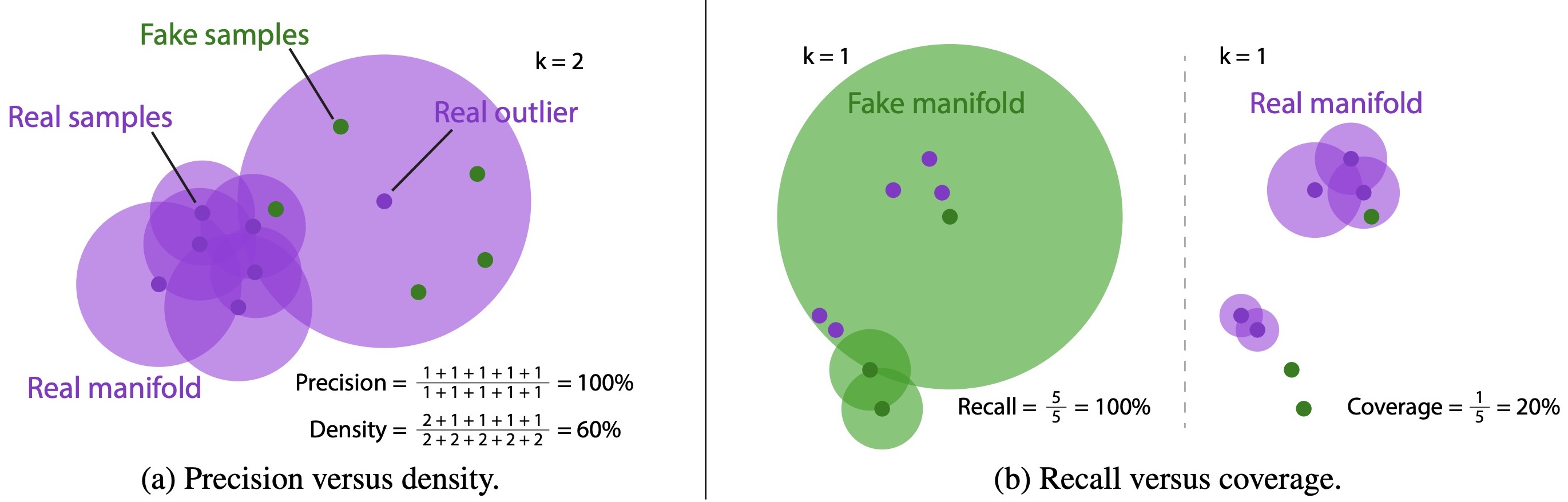}
\caption{Pictorial depiction of density and coverage measures~\citep{naeem2020reliable} (See text for definitions). Note that in the recall versus coverage figure (panel b), the real and fake samples are identical across left and right. a) Here, the real manifold is overestimated due to the real outlier sample. Generating many fake samples around the real outlier increases the precision measure. b) Here, although the fake samples are far from the modes in real samples, the recall is perfect because of the overestimated fake manifold. 
Figure compiled from~\cite{naeem2020reliable}.}
\label{fig:naeem2020reliable}
\end{center}
\end{figure}

\cite{naeem2020reliable} argue that even the latest version of the precision and recall metrics are still not reliable as they a) fail to detect the match between two identical distributions, b) are not robust against outliers, and c) have arbitrarily selected evaluation hyperparameters. To solve these issues, they propose density and coverage metrics. Precision counts the binary decision of whether the fake data $Y_j$ is contained in any neighbourhood sphere. Density, instead, counts how many real-sample neighbourhood spheres contain $Y_j$. See Fig.~\ref{fig:naeem2020reliable}. They analytically and experimentally show that density and coverage provide more interpretable and reliable signals for practitioners than the existing measures\footnote{\url{https://github.com/clovaai/generative-evaluation-prdc}}. 


\subsubsection{Alpha Precision and Recall}
\begin{figure}[t]
\begin{center}
   \includegraphics[width=.6\linewidth]{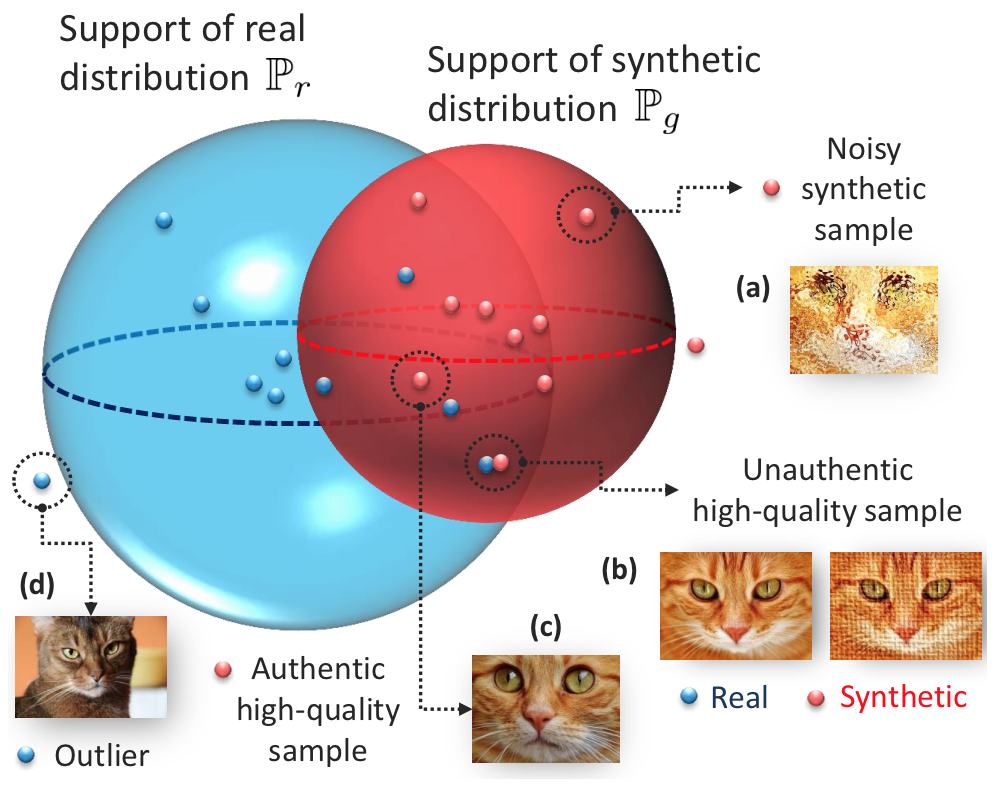}
   \caption{Illustration of $\alpha$-Precision, $\beta$-Recall and
Authenticity metrics~\citep{alaa2021faithful}. Blue and red spheres correspond to the $\alpha$ and $\beta$-supports of real and generative distributions, respectively.
Blue and red points correspond to real and synthetic data. (a) Generated samples falling outside the blue sphere look unrealistic or noisy. (b) Overfitted models can generate high-quality samples that are “unauthentic” because they are copied from training data. (c) High-quality samples should reside in the blue sphere.
(d) Outliers do not count in the $\beta$-Recall metric. (Here, $\alpha$=$\beta$=0.9,
$\alpha$-Precision=8/9, $\beta$-Recall=4/9, Authenticity=9/10). Figure from~\cite{alaa2021faithful}.}
\label{fig:Alaa}
\end{center}
\end{figure}

\cite{alaa2021faithful} introduce a 3-dimensional evaluation metric, ($\alpha$-Precision, $\beta$-Recall, Authenticity), to characterizes the fidelity, diversity and generalization power of generative models (Fig.~\ref{fig:Alaa}). The first two assume that a fraction 1 − $\alpha$ (or 1 − $\beta$) of the real (and synthetic) data are ``outliers'', and $\alpha$ (or $\beta$) are ``typical''.
$\alpha$-Precision is the fraction of synthetic samples that resemble the ``most typical'' $\alpha$ real samples, whereas $\beta$-Recall is the fraction of real samples covered by the most typical $\beta$ synthetic samples. $\alpha$-Precision and $\beta$-Recall are evaluated for all $\alpha$, $\beta$ $\in [0, 1]$, providing entire precision and recall curves instead of a single number. To compute both metrics, the (real and synthetic) data are embedded into hyperspheres with most samples concentrated around the centers.
Typical samples are located near the centers whereas outliers are close to the boundaries. To quantify Generalization they introduce the Authenticity metric to measure the probability that a synthetic sample is copied from the training data. Some other works that have proposed extensions to P\&R include~\cite{djolonga2020precision,simon2019revisiting,kynkaanniemi2019improved}.

\subsection{Duality GAP Metric}
\cite{grnarova2019domain} leverage the notion of duality gap from game theory to propose a domain-agnostic and computationally-efficient measure that can be used to assess different models as well as monitoring the progress of a single model throughout training. Intuitively, duality gap measures the 
sub-optimality (\emph{w.r.t.} an equilibrium) of a given solution (G,D) where G and D and the generator and the discriminator, respectively. 
They also show that their measure highly correlates with FID on natural image datasets, and can also be used in other modalities such as text and sound. Further, their measure requires no labels or a pretrained classifier, making it domain agnostic. 
In a follow-up work, \cite{sidheekh2021characterizing} extend the notion
of duality gap to proximal duality gap that is applicable to the general context of training GANs.

\subsection{Spectral Methods}
\begin{figure}[t]
\begin{center}
\includegraphics[width=1\linewidth]{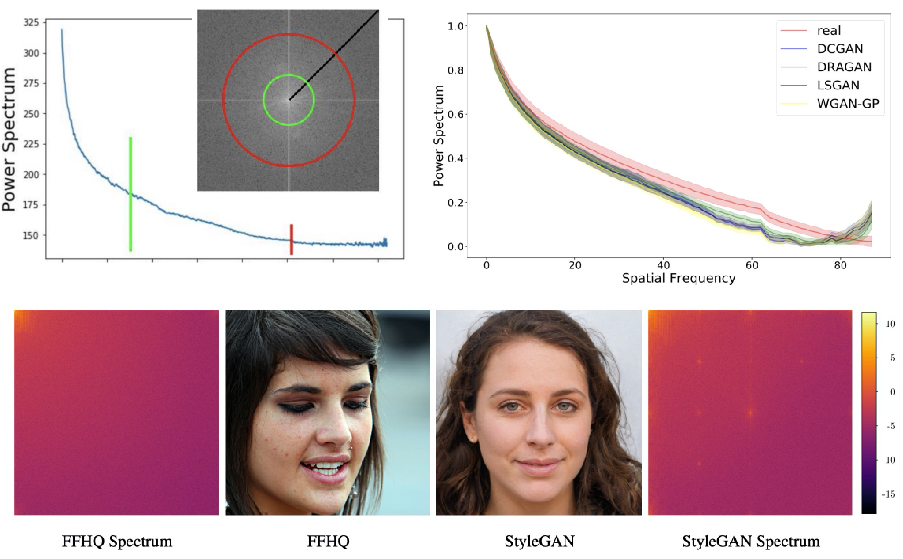}
\caption{Top-left) An example of azimuthal integral for an image (inset) taken from~\cite{durall2020watch}. In the 1D Power Spectrum, each frequency component is the radial integral over the 2D spectrum (shown in red and green). Top-right) Statistics (mean and variance) after azimuthal integration over the power-spectrum of real and GAN generated images over CelebA dataset~\citep{liu2015deep}. Notice in particular the sharp increase at higher frequencies. 
Bottom) A side-by-side comparison of real and generated faces in spatial and frequency domains. The left-most panel shows that mean DCT spectrum of the FFHQ with a sample from this dataset next to it. The right-most side shows he mean DCT spectrum of a dataset sampled from StyleGAN trained on FFHQ, with a generated face to its left. Results are averaged over 10K images. Image from~\cite{frank2020leveraging}. }
\label{fig:spectral}
\end{center}
\end{figure}

A number of studies (\eg~\cite{durall2020watch,frank2020leveraging,dzanic2019fourier,zeng2017statistics}) have shown that current generators are unable to correctly approximate the spectral distributions of real data. 
\cite{durall2020watch} showed that up-scaling operations commonly used in GANs (\eg up-convolutions) alter the spectral properties of the images causing high frequency distortions in the output. 
This can be seen in the average azimuthal integration of the power spectrum shown in the top panel of Fig.~\ref{fig:spectral}. 
Based on this observation, they proposed a novel spectral regularization term to compensate spectral distortions. They also proposed a very simple but highly accurate detector for generated images and videos, \ie a DeepFake detector.

\subsection{Caption Score (CapS)}
Almost all existing metrics are solely for evaluating models that generative image. They do not take into account the corresponding text in the context of text-to-image generation. Thus, they may be fooled by a network that ignores the textual input and only focuses on generating realistic looking images.
To remedy this shortcoming, \cite{ding2021cogview} introduce the caption score to evaluate the correspondence between images and text. This score measures the quality and accuracy for text-image generation at a finer granularity than FID and Inception Score (IS), and is defined as:
\[
\text{CapS}(x,t)=\sqrt[|t|]{\Pi_{i=0}^{|t|} p(t_i|x, t_{0:i-1}),} 
\]
where $t$ is a sequence of text tokens and $x$ is the image. The higher the CapS, the better.
In addition to CapS, they also report results using other measures such as FID and IS.

\subsection{Perplexity}
A commonly used metric to evaluate generative models of text is the perplexity.
It measures the probability for a sentence to be produced by a language model that has been trained on a dataset: 


$\text{perplexity }=\prod_{t=1}^{T}\left(\frac{1}{P_{\mathrm{LM}}\left(\boldsymbol{x}^{(t+1)} \mid \boldsymbol{x}^{(t)}, \ldots, \boldsymbol{x}^{(1)}\right)}\right)^{1 / T}
$

\noindent where $\boldsymbol{x}$ represent a single word and $T$ is the sentence length. 
The lower the perplexity value, the better the model (averaged over a set of test sentences). See~\href{https://towardsdatascience.com/how-to-evaluate-text-generation-models-metrics-for-automatic-evaluation-of-nlp-models-e1c251b04ec1}{here}, \href{https://en.wikipedia.org/wiki/Perplexity}{here},~\cite{iqbal2020survey}, and~\cite{tevet2018evaluating} for some other measures for evaluating text generation models.






\section{New Qualitative GAN Evaluation Measures}
A number of new qualitative measures have also been proposed. These measures typically focus on how convincing a generated image is from human perception perspective. 

\subsection{Human Eye Perceptual Evaluation (HYPE)}
\begin{figure}[t]
\begin{center}
\includegraphics[width=\linewidth]{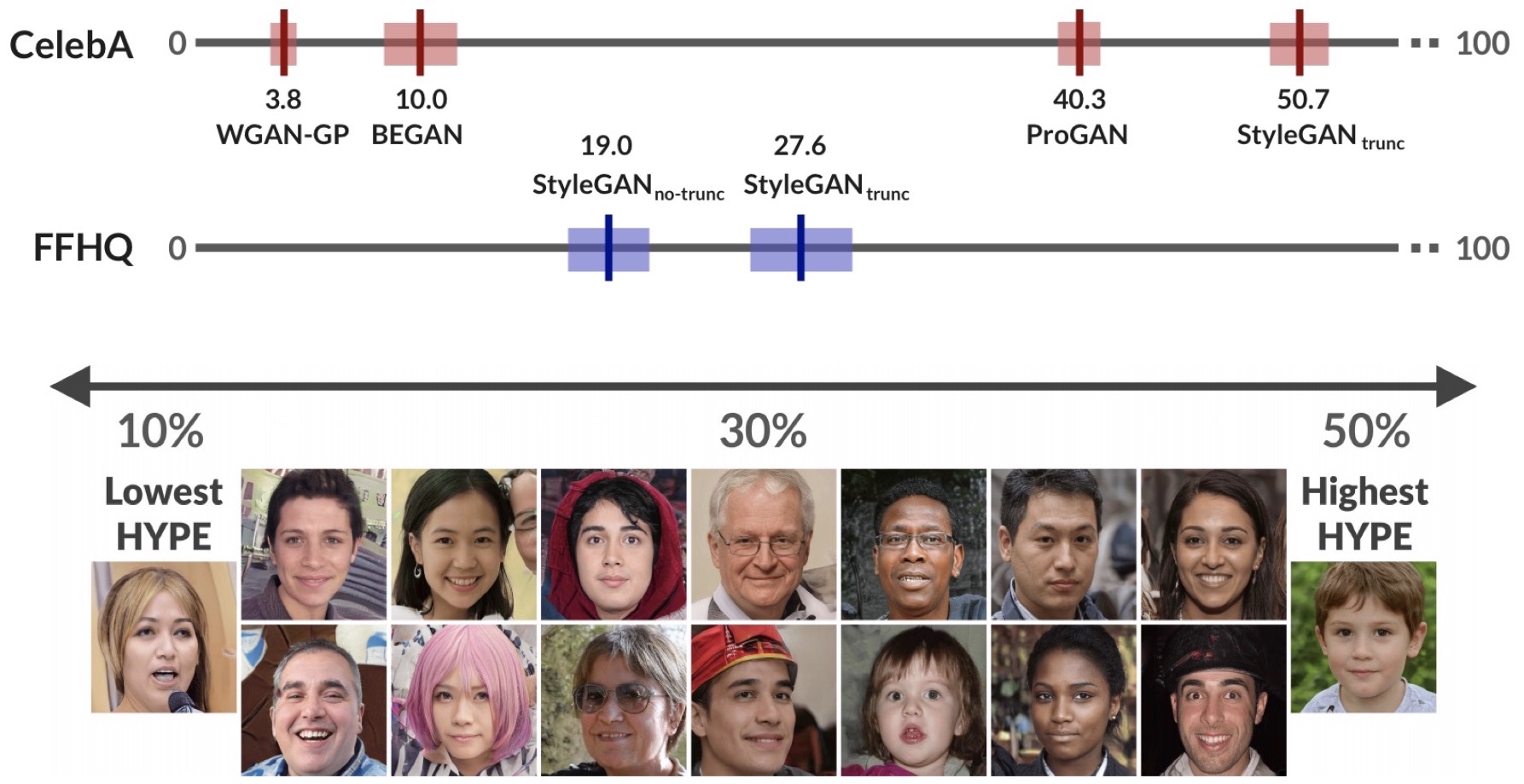}
\caption{HYPE tests generative models for how realistic their images look to the human eye~\citep{zhou2019hype}. Top) HYPE scores of different models over CelebA and FFHQ datasets. A score of 50\% represents indistinguishable results from real, while a score above 50\% represents hyper-realism. Bottom) Example images sampled with the truncation trick from StyleGAN trained on FFHQ dataset. Images on the right have the highest HYPE scores (\ie exhibit the highest perceptual fidelity). Figure compiled from~\cite{zhou2019hype}.}
\label{fig:zhou2019hype}
\end{center}
\end{figure}

\cite{zhou2019hype} propose a human-in-the-loop evaluation approach that is grounded in psychophysics research on visual perception (Fig.~\ref{fig:zhou2019hype}). 
They introduce two variants. The first one measures visual perception under adaptive time constraints to determine the threshold at which a model's outputs appear real (\eg 250 ms). The second variant is less expensive and measures human error rate on fake and real images without time constraints\footnote{An important point when conducting experiments of this sort (\eg~\cite{denton2015deep}) is to make sure that real and generated images are  shuffled during the presentation. Otherwise, subjects maybe able to guess what type of images are shown in a session from the first few images, if all the images in the session in either real or fake.}. \cite{kolchinski2019approximating} propose an approach to approximate HYPE and report 66\% accuracy in predicting human scores of image realism, matching the human inter-rater agreement rate. A major drawback with human evaluation approaches such as HYPE is scaling them. One way to remedy this is to train a model from human judgments and interact with a person only when the model is not certain. HYPE is more reliable compared to automated ones but cannot be used for monitoring the training process.

\subsection{Neuroscore}
\begin{figure}[t]
\begin{center}
\includegraphics[width=1\linewidth]{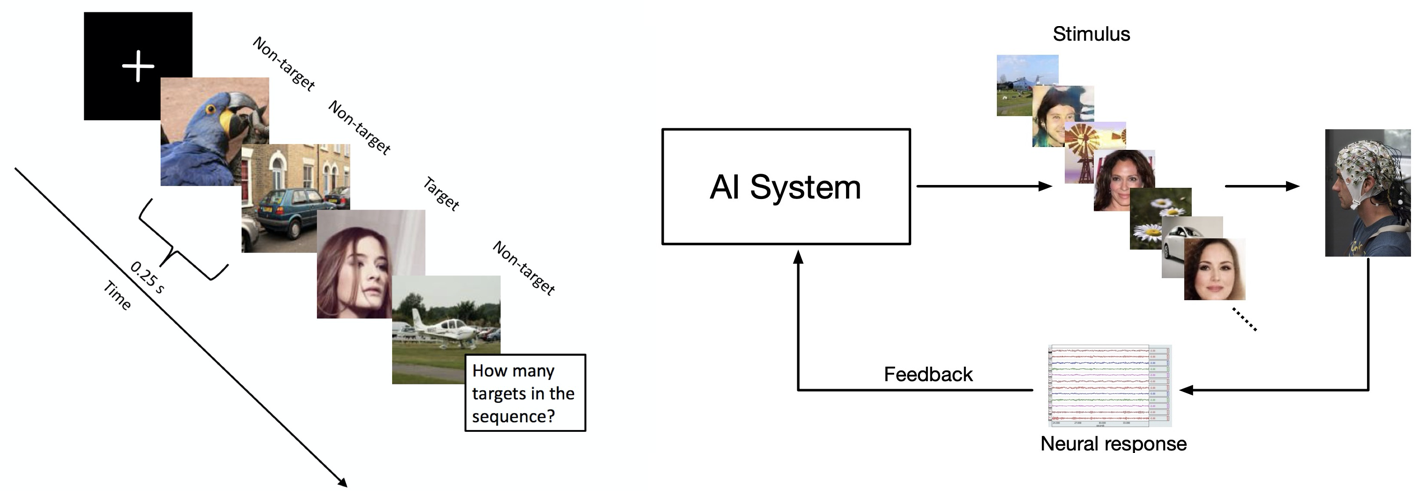}
\caption{
Left) An example of RSVP experimental protocol in which a rapid image stream (4 images per second) containing target and non-target images is presented to participants. Participants are instructed to search for real face images. The idea is that GAN generated faces will elicit a different response than real faces. Right) Schematic diagram of neuro-AI interface for computing the Neuroscore. Generated images by a GAN are shown to the participants and the corresponding recorded neural responses are used to evaluate the performance. Figure compiled from~\cite{wang2020use}.}
\label{fig:wang2020use}
\end{center}
\end{figure}

\cite{wang2020use} outline a method called Neuroscore using neural signals and rapid serial visual presentation (RSVP) to directly measure human perceptual response to generated stimuli. Participants are instructed to attend to target images (real and generated) amongst a larger set of non-target images (left panel in Fig.~\ref{fig:wang2020use}). This paradigm, known as the oddball paradigm, is commonly used to elicit the P300 event-related potential (ERP), which is a positive voltage deflection typically occurring between 300 ms and 600 ms after the appearance of a rare visual target. Fig.~\ref{fig:wang2020use} shows a depiction of this approach. Wang \etal~show that Neuroscore is more consistent with human judgments compared to the conventional metrics. They also trained a convolutional neural network to predict Neuroscore from GAN-generated images directly without the need for neural responses.

\subsection{Seeing What a GAN Can Not Generate}
\begin{figure}[t]
\begin{center}
   \includegraphics[width=1\linewidth]{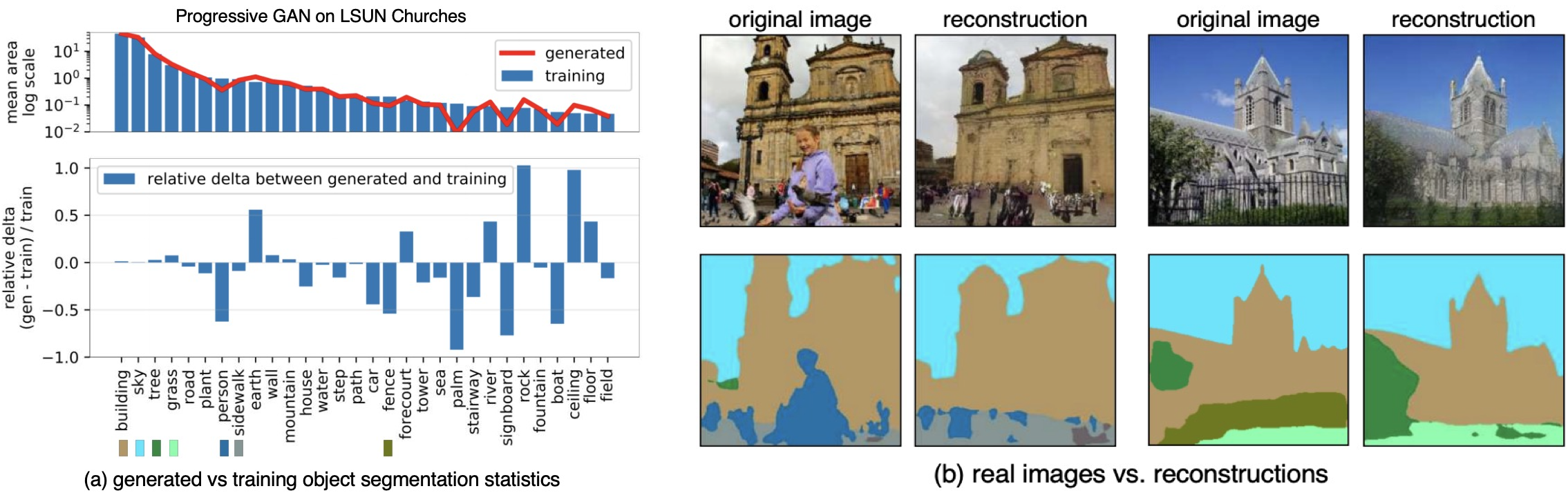}
   \caption{Seeing what a GAN cannot generate~\citep{bau2019seeing}. (a) Distribution of object segmentations in the training set of LSUN churches vs. the corresponding distribution over the generated images. Objects such as people, cars, and fences are dropped by the generator. (b) Pairs of real images and their reconstructions in which individual instances of a person and a fence cannot be generated. 
   Figure compiled from~\cite{bau2019seeing}.}
\label{fig:bau2019seeing}   
\end{center}
\end{figure}

\cite{bau2019seeing} visualize mode collapse at both distribution level and instance level. They employ a semantic segmentation network to compare the distribution of segmented objects in the generated images versus the real images. 
Differences in statistics reveal object classes that are omitted by a GAN. 
Their approach allows to visualize the GAN’s omissions for an omitted class and to compare differences between individual photos and their approximate inversions by a GAN. Fig.~\ref{fig:bau2019seeing} illustrates this approach.

\subsection{Measuring GAN Steerability} 
\begin{figure}[t]
\begin{center}
    \includegraphics[width=1\linewidth]{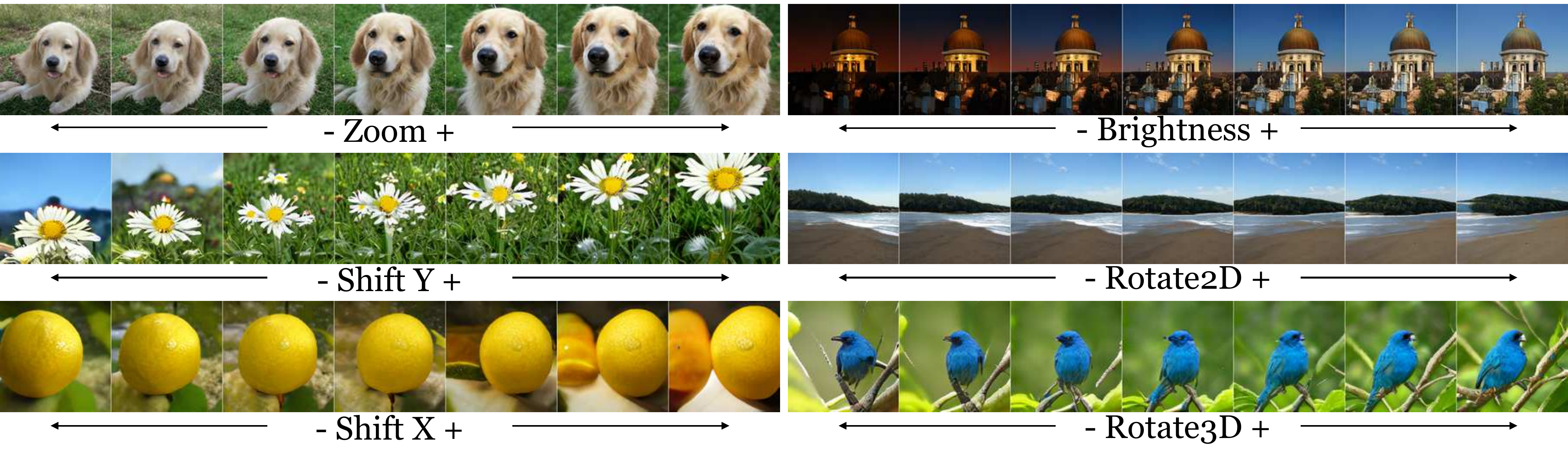}
   \includegraphics[width=1\linewidth]{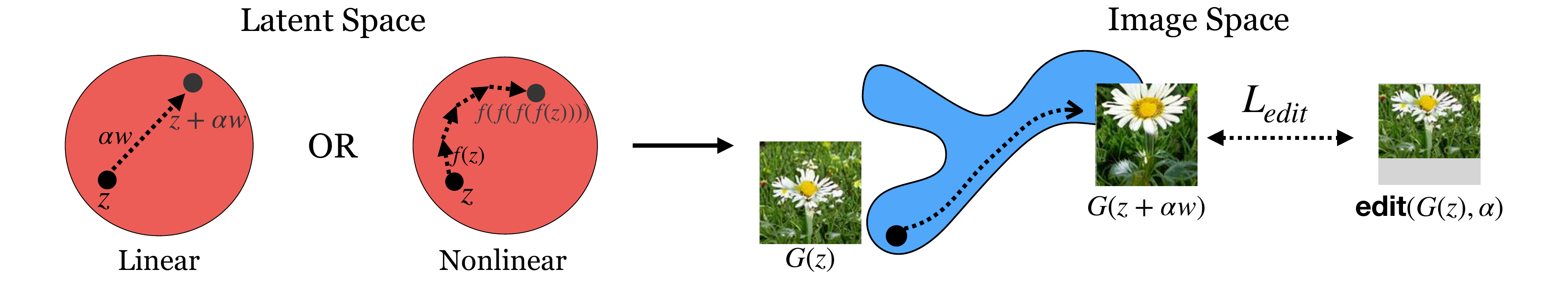}
   \caption{Top) A walk in the latent space of a GAN corresponds to visual transformations such as zoom and camera shift,
Bottom) The goal is to find a path in $z$ space (linear or non-linear) to transform the generated image $G(z)$ to its edited version $\texttt{edit}(G(z,\alpha))$, \eg an $\alpha\times$ zoom. 
To measure steerability, the distributions of a given attribute in real images and generated images (after walking in the latent space) are compared.
Figure from~\cite{jahanian2019steerability}.}
\label{fig:jahanian2019steerability}
\end{center}
\end{figure}

\cite{jahanian2019steerability} propose a method to quantify the degree to which basic visual transformations are achieved by navigating the latent space of a GAN (See Fig.~\ref{fig:jahanian2019steerability}). They first learn a $N$-dimensional vector representing the optimal path in the latent space for a given transformation. Formally, the task it to learn the walk $w$ by minimizing the following objective function: 
\[
 w^* =  \texttt{argmin}_w \ \mathbb{E}_{z,\alpha} [\mathcal{L} ( G(z\!+\!\alpha w), \texttt{edit}(G(z), \alpha) )],
\]
where $\alpha$ is the step size, and $\mathcal{L}$ is the distance between the generated image $G(z + \alpha w)$ after taking an $\alpha$-step in the latent direction and the target image \texttt{edit}($G(z)$, $\alpha$). The latter is the new image derived from the source image $G(z)$. To quantify how well a desired image manipulation under each transformation is achieved, the distributions of a given attribute (\eg ``luminance'') in the real data and generated images (after walking in latent space) are compared.

\subsection{GAN Dissection}

\begin{figure}[htbp]
\begin{center}
\includegraphics[width=\linewidth]{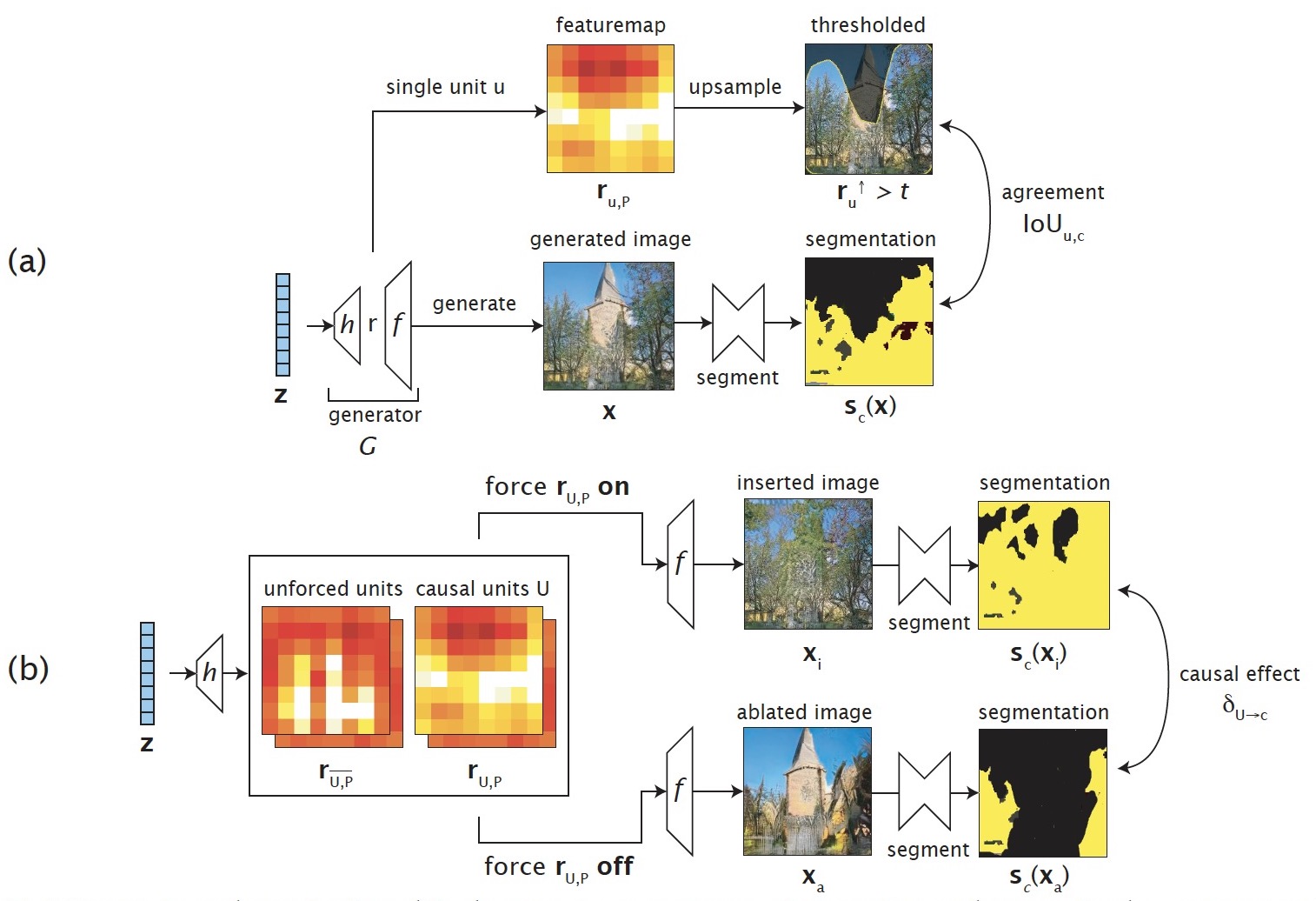}
\includegraphics[width=.9\linewidth]{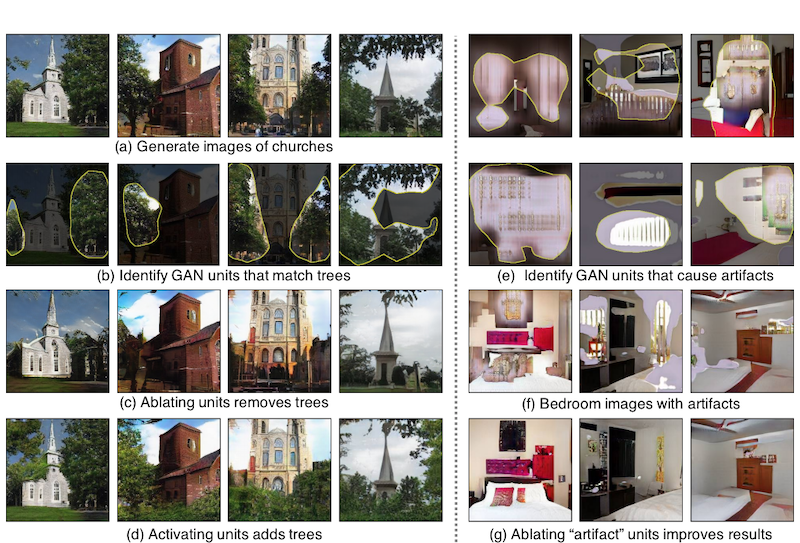}
\caption{An overview of the GAN dissection approach~\citep{bau2018gan}. 
Top) Measuring the relationship between representation units and trees in the output using (a) dissection and (b) intervention. Dissection measures agreement between a unit $u$ and a concept $c$ by comparing its thresholded upsampled heatmap with a semantic segmentation of the generated image $s_c(x)$. Intervention measures the causal effect of a set of units $U$ on a concept $c$ by comparing the effect of forcing these units on (unit insertion) or off (unit ablation). The segmentation $s_c$ reveals that trees increase after insertion and decrease after ablation. The average difference in the tree pixels measures the average causal effect. Bottom) Applying the dissection method to a generated outdoor church image. Dissection method can also be used to diagnose and improve GANs by identifying and ablating the artifact-causing units (panels e to g). Figure compiled from~\cite{bau2018gan}.}
\label{fig:bau2018gan}
\end{center}
\end{figure}

\cite{bau2018gan} propose a technique to dissect and visualize the inner workings of an image generator\footnote{A demo of this work is available at~\href{https://www.google.com/search?q=GAN+dissection&oq=GAN+dissection&aqs=chrome..69i57.7356j0j1&sourceid=chrome&ie=UTF-8}{here}.}. The main idea is to identify GAN units (\ie generator neurons) that are responsible for semantic concepts and objects (such as tree, sky, and clouds) in the generated images. 
Having this level of granularity into the neurons allows editing existing images (\eg to add or remove trees as shown in Fig.~\ref{fig:bau2018gan}) by forcefully activating and deactivating (ablating) the corresponding units for the desired objects. Their technique also allows finding artifacts in the generated images and hence can be used to evaluate and improve GANs. A similar approach has been proposed in~\cite{park2019semantic} for semantic manipulation and editing of GAN generated images\footnote{See~\href{https://blogs.nvidia.com/blog/2019/03/18/gaugan-photorealistic-landscapes-nvidia-research/}{here} for an illustration.}.

\subsection{A Universal Fake {\em vs.} Real Detector}
\begin{figure}[t]
\begin{center}
   \includegraphics[width=1\linewidth]{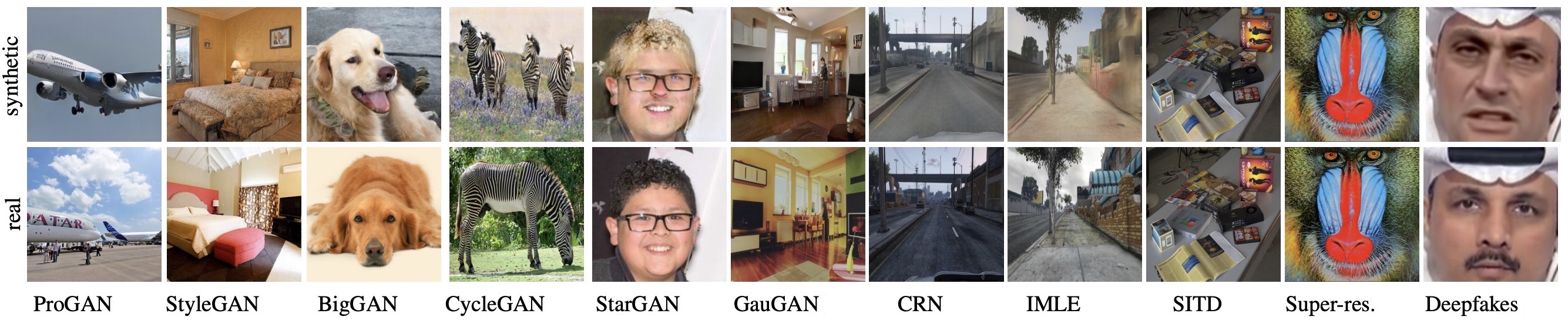}
   \caption{\cite{wang2020cnn} show that a classifier trained to distinguish images generated by only one GAN (ProGAN, the left-most column) from real ones can detect the images generated by other generative models (remaining columns) as well. Please see \url{https://peterwang512.github.io/CNNDetection/}. }
\label{fig:wang2020cnn}
\end{center}
\end{figure}

\cite{wang2020cnn} ask whether it is possible to create a ``universal'' detector to distinguish real images from synthetic ones using a dataset of synthetic images generated by 11 CNN-based generative models. See Fig.~\ref{fig:wang2020cnn}.
With careful pre- and post-processing and data augmentation, they show that an image classifier trained on only one specific CNN generator is able to generalize well to unseen architectures, datasets, and training methods. They highlight that today’s CNN-generated images share common systematic flaws, preventing them from achieving realistic image generation. Similar works have been reported in~\cite{chai2020makes,yu2019attributing}. In alignment with these results,
\cite{gragnaniello2021gan} also conclude that we are still far from having reliable tools for GAN image detection.

\section{Discussion}

\subsection{GAN Benchmarks and Analysis Studies}
\begin{figure}[t]
\begin{center}
\includegraphics[width=1\linewidth]{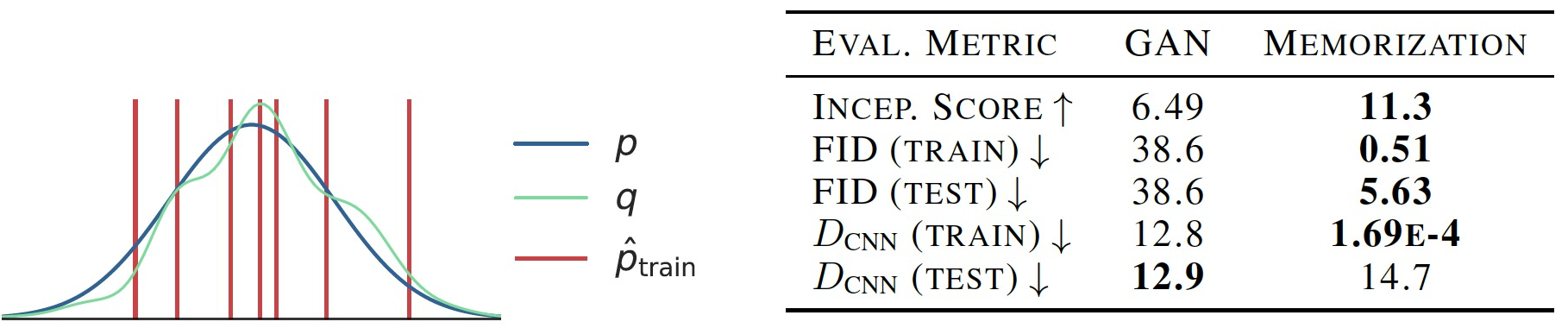}
\caption{Left) \cite{gulrajani2020towards} show that common evaluation measures such as IS and FID prefer a model that memorizes the dataset ($p_{train}$, red) to a model ($q$, green) which imperfectly fits
the true distribution ($p$, blue) but covers
more of $p$’s support. Right) Neural network divergence, DCNN, discounts memorization and prefers GANs that generalize beyond the training set. Figure compiled from~\cite{gulrajani2020towards}}
\label{fig:gulrajani2020towards}
\end{center}
\end{figure}

Following previous works (\eg~\cite{lucic2017gans,shmelkov2018good}), new studies have investigated formulating good criteria for GAN evaluation, or have conducted systematic GAN benchmarks. \cite{gulrajani2020towards} argue that a good evaluation measure should not have a trivial solution (\eg memorizing the dataset) and show that many scores such as IS and FID can be won by simply memorizing the training data (Fig.~\ref{fig:gulrajani2020towards}). They suggest that a necessary condition for a metric not to behave this way is to have a large number of samples. They also propose a measure based on neural network divergences (NND). NND works by training a discriminative model to discriminate between samples of the generative model and samples from a held-out test set. The poorer the discriminative model performs, the better the generative model is.
Through experimental validation, they show that NND can effectively measure diversity, sample quality, and generalization. \cite{lee2020mimicry} introduce Mimicry\footnote{\url{https://github.com/kwotsin/mimicry}}, a lightweight PyTorch library that provides implementations of popular GANs and evaluation metrics to closely reproduce reported scores in the literature. They also compare several GANs on seven widely-used datasets by training them under the same conditions, and evaluating them using three popular GAN metrics. Some tutorials and interaction tools have also been developed to understand, visualize, and evaluate GANs\footnote{See \href{https://towardsdatascience.com/graduating-in-gans-going-from-understanding-generative-adversarial-networks-to-running-your-own-39804c283399}{this link} and \url{https://poloclub.github.io/ganlab/}.}.

Also mention that deepfakes might be detected because they have different spectral properties than real images. See for example~\cite{durall2020watch}

\subsection{Assessing Fairness and Bias of Generative Models}
Fairness and bias in ML algorithms, datasets, and commercial products have become growing concerns recently (\eg~\cite{buolamwini2018gender}) and have attracted widespread attention from public and media\footnote{\url{https://thegradient.pub/pulse-lessons/}}. 
Even without a single precise definition for fairness~\citep{verma2018fairness}, it is still possible to observe the lack of fairness across many domains and models, with GANs being no exception. Some recent works (\eg~\cite{xu2018fairgan,yu2020inclusive}
have tried to mitigate the bias in GANs or use GANs to reduce bias in other ML algorithms (\eg~\cite{sattigeri2019fairness,mcduff2019characterizing}).
Bias can enter a model during training (through data, labeling, architecture), evaluation (such as who created the evaluation measure), as well as deployment (how is the model being distributed and for what purposes). Therefore it is critical to address these aspects when evaluating and comparing generative models. 

\subsection{Connection to Deepfakes}
\begin{figure}[t]
\begin{center}
   \includegraphics[width=.45\linewidth]{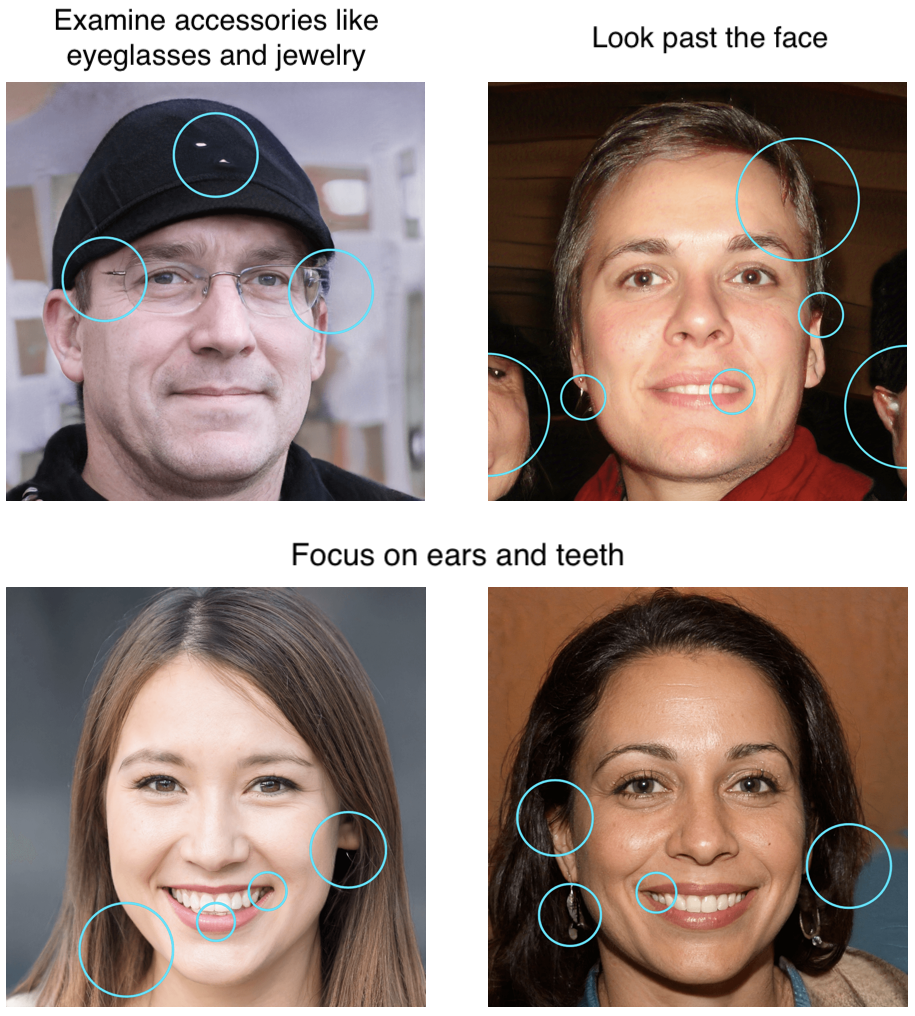}
  \hspace{15pt}
   \includegraphics[width=.45\linewidth]{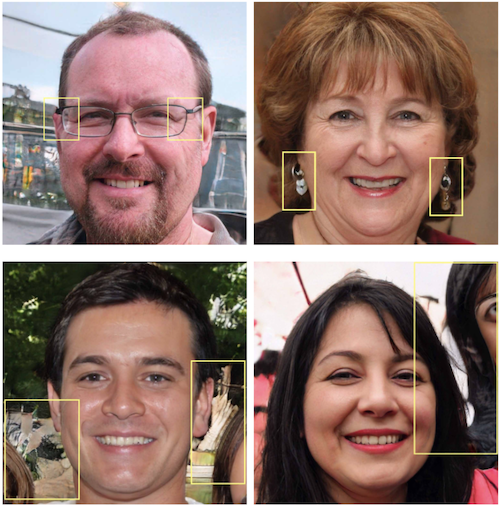}
   \caption{Sample generated/fake faces and cues to tell them apart from real ones. See also~\url{https://chail.github.io/patch-forensics/}. Image courtesy of Twitter.}
\label{fig:fake1}   
\end{center}
\end{figure}

\begin{figure}[htbp]
\begin{center}
   \includegraphics[width=.8\linewidth]{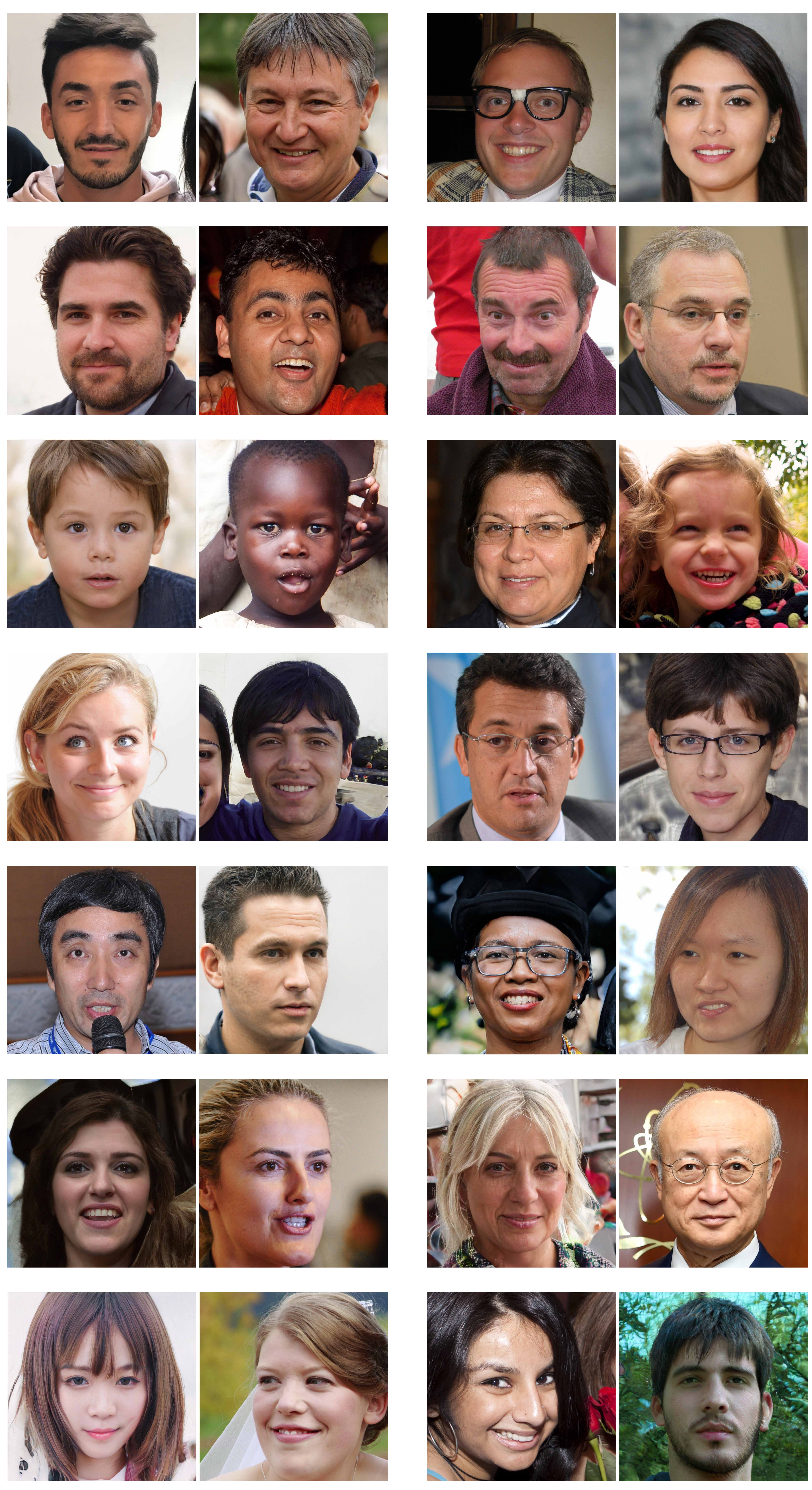}
   \caption{Can you determine which face in each pair is real? Key (L for left and R for right) row wise: L, L, R, L, R, R, L, L, L, R, R, R, R, L. Images taken from~\url{https://www.whichfaceisreal.com/}.}
\label{fig:fake2}   
\end{center}
\end{figure}

\begin{figure}[htbp]
\begin{center}
   \includegraphics[width=.7\linewidth]{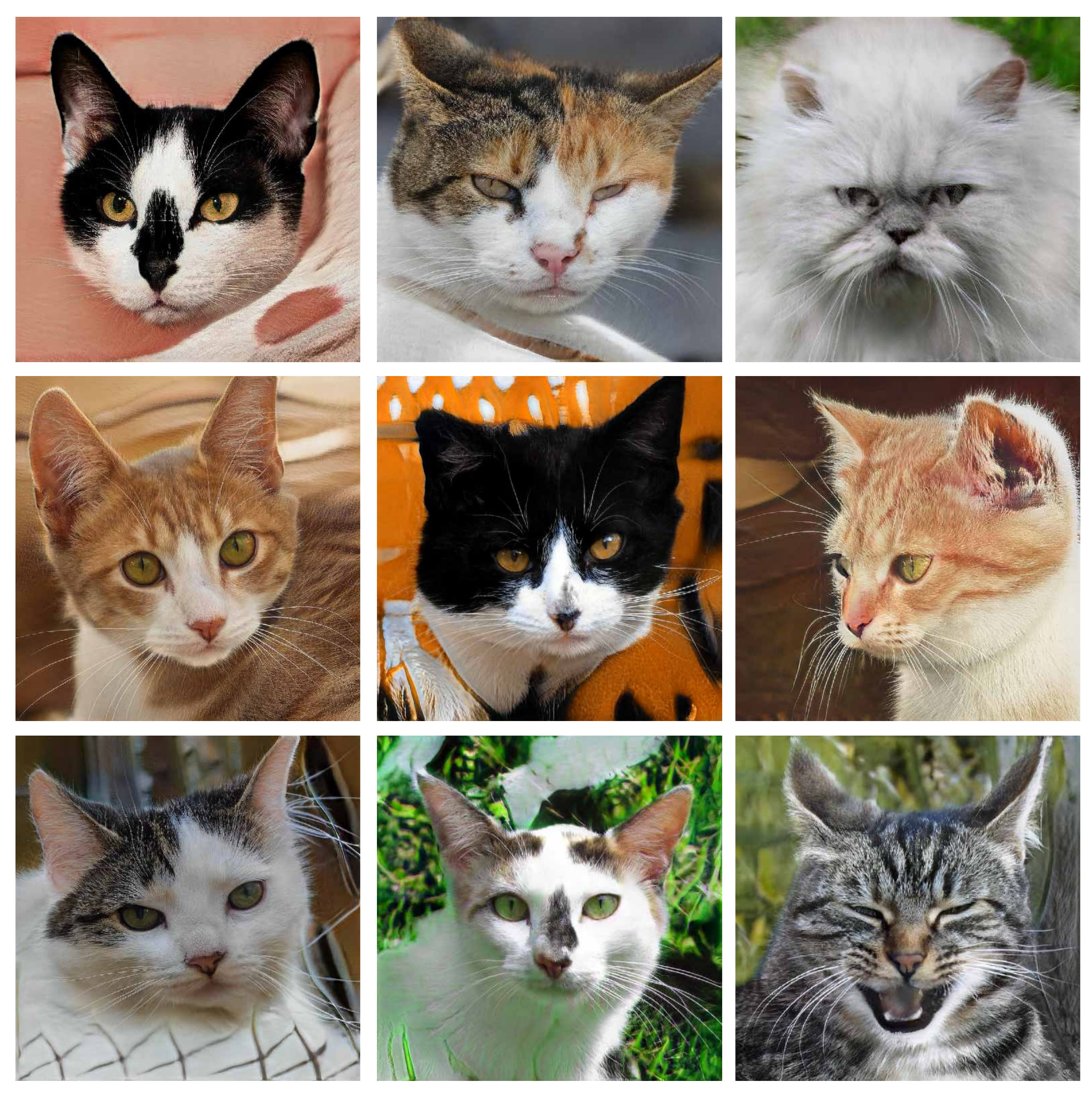}
   \includegraphics[width=.7\linewidth]{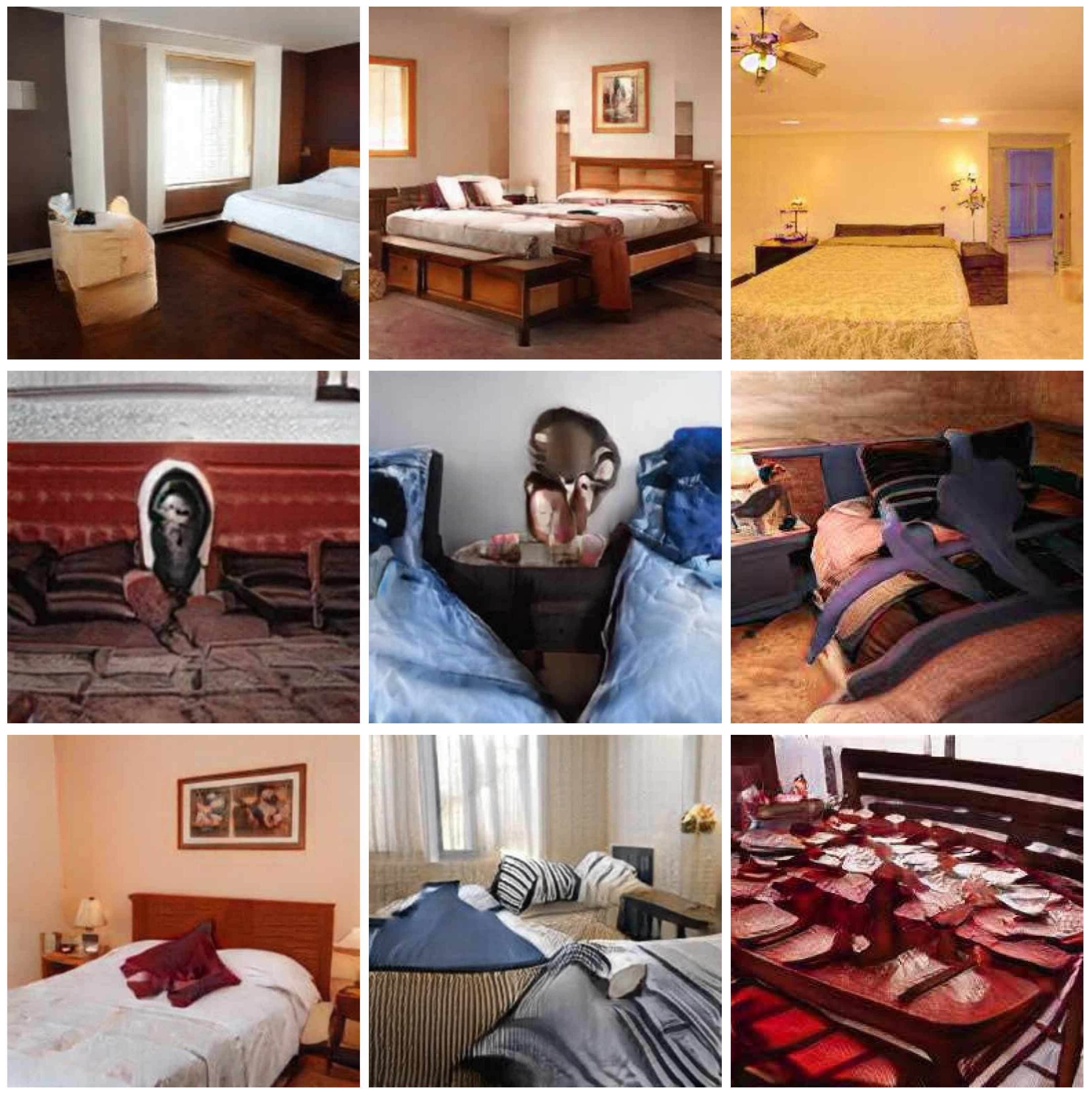}   
   \caption{Some sample generated cats (top) and beds (bottom), generated from \url{http://thiscatdoesnotexist.com/} and \url{https://thisrentaldoesnotexist.com/}, respectively. See also \url{https://thisxdoesnotexist.com/}. Although images look realistic in the first glance, a closer examination reveals the artifacts (See it for yourself!).}
\label{fig:fake3}   
\end{center}
\end{figure}

A alarming application of generative models is fabricating fake content\footnote{Some concerns can be found at these links \href{https://www.skynettoday.com/overviews/state-of-deepfakes-2020}{1},  \href{https://www.cnn.com/interactive/2020/10/us/manipulated-media-tech-fake-news-trnd/}{2}, \href{https://www.cnn.com/2019/02/28/tech/ai-fake-faces/index.html}{3}, and \href{https://www.theguardian.com/technology/2020/jan/13/what-are-deepfakes-and-how-can-you-spot-them}{4}}. It is thus crucial to develop tools and techniques to detect and limit their use (See~\cite{tolosana2020deepfakes} for a survey on deepfakes). There is a natural connection between deepfake detection and GAN evaluation. Obviously, as generative models improve, it becomes increasingly harder for a human to distinguish between what is real and what is fake (manipulated images, videos, text, and audio). Telling the degree of fakeness of an image or video directly tells us about the performance of a generator. Even over faces, where generative models excel, it is still possible to detect fake images (Fig.~\ref{fig:fake1}), although in some cases it can be very daunting (Fig.~\ref{fig:fake2}). Difficulty of deepfake detection for humans is also category dependent, with some categories such as faces, cats, and dogs being easier than bedrooms or cluttered scenes (Fig.~\ref{fig:fake3}). Fortunately, or unfortunately, it is still possible to build deep networks that can detect the subtle artifacts in the doctored images (\eg using the universal detectors mentioned above~\citep{wang2020cnn,chai2020makes}). Moving forward, it is important to study whether and how GAN evaluation measures can help us mitigate the threat from deepfakes.

\section{Summary and Conclusion}
\begin{figure}[htbp]
\begin{center}
   \includegraphics[width=\linewidth]{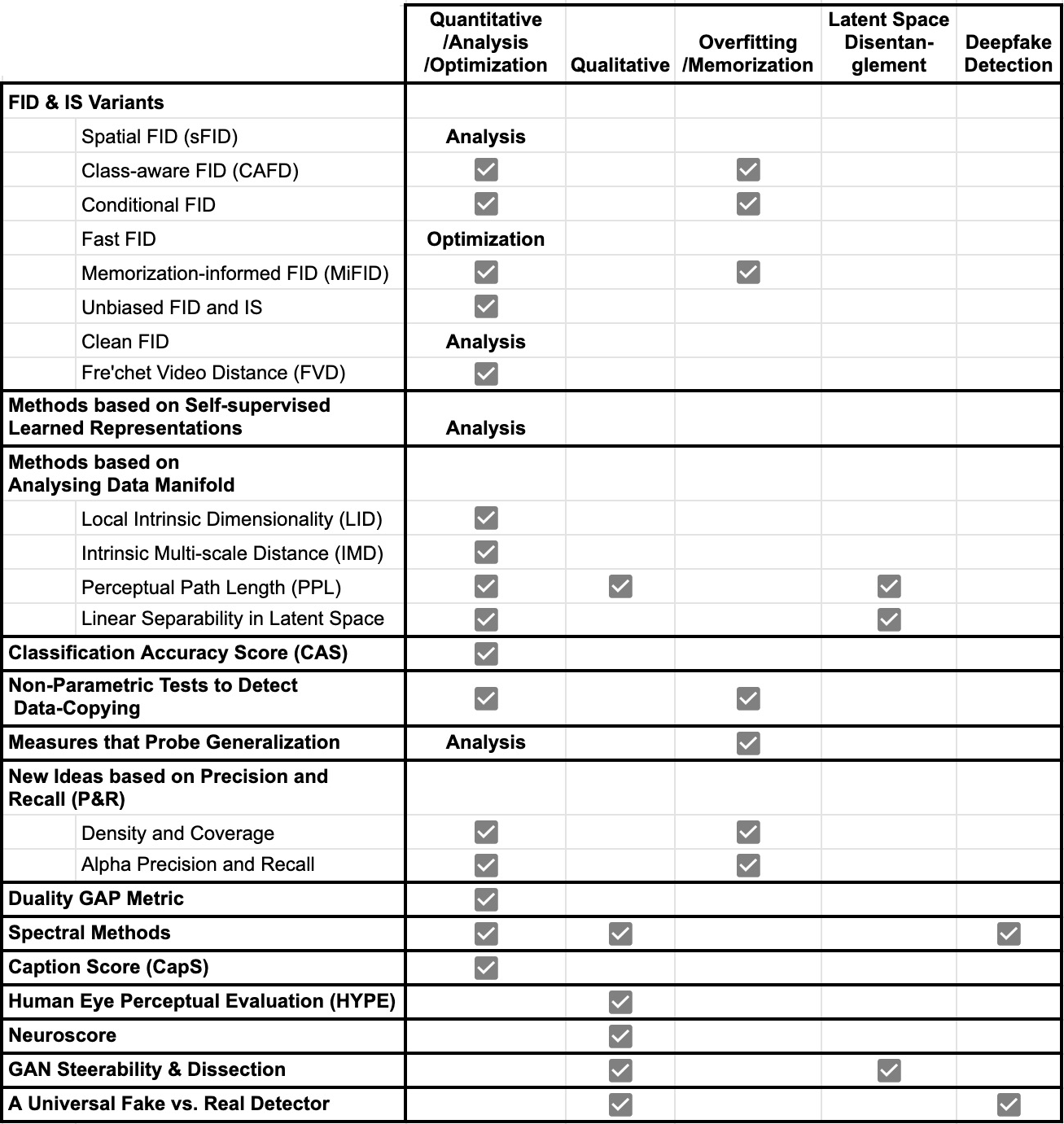}
   \caption{A summary of evaluation measures covered in this work.}
\label{fig:table}   
\end{center}
\end{figure}

Here, I reviewed a number of recently proposed GAN evaluation measures. A summary of the discussed measures is shown in Figure~\ref{fig:table}. Similar to generative models, evaluation measures are also evolving and improving over time. Although some measures such as IS, FID, P\&R, and PPL are relatively more popular than others, objective and comprehensive evaluation of generative models is still an open problem. Some directions for future research in GAN evaluation are as follows. 

\begin{enumerate}
    \item Prior research has been focused on examining generative models of faces or scenes containing one or few objects. Relatively, less effort has been devoted to assess how good GANs are in generating more complex scenes such as bedrooms, street scenes, and nature scenes. As an example, \cite{casanova2020generating} is an early effort in this direction. 
    \item Generative models have been primarily evaluated in terms of the quality and diversity of their generated images. Other dimensions such as generalization and fairness have been less explored. Generalization assessment can provide a deeper look into what generative models learn. For example, it can tell how and to what degree models capture compositionality (\eg does a model generate the right number of paws, nose, eyes, etc for dogs?) and logic (does a model properly capture physical properties such as gravity, light direction, reflection, and shadow?). Evaluating models in terms of fairness is critical for mitigating the potential risks that may arise at the deployment time and to ensure that a model has the right societal impact. 
    \item An important matter in GAN evaluation is task dependency. In other words, how well a generative model works depends on its intended use. Sometimes, it might be easier to evaluate a model on downstream tasks when those tasks have a clear target for a given input. In some tasks (\eg graphics applications such as image synthesis, image translation, image inpainting, and attribute manipulation) image quality is more important, whereas in some other tasks (\eg generating synthetic data for data augmentation) diversity may weigh more. Thus, evaluation metrics should be tailored to the target task. It should be note that the representation power of the discriminator or encoder of a GAN does not necessarily reflect its sample quality and diversity.
\item Having good evaluation measures is important not only for ranking generative models but also for diagnosing their errors. Reliable GAN evaluation is specially important in domains where humans are less attuned to discern the quality of samples (\eg medical images). An important question thus is whether current evaluation measures generalize across different domains (\ie are domain-agnostic). 
\item The degree to which generative models memorize the training data is still unclear. 
A common technique to assess memorization is to look for nearest neighbors. It is known that this approach has several shortcomings~\citep{theis2015note,borji2019pros}. Motivated by works that study memorization in supervised learning works,~\cite{van2021memorization} proposed new methods to understand and quantify memorization in generative models.
\item Ultimately, future works should also study how research in evaluating performance of generative models can help mitigate the threat from fabricated content. In this regard, extending works such as~\cite{durall2020watch,wang2020cnn}, conducting benchmarks on standard datasets containing fabricated content, as well as cross-talks between research in GAN evaluation and deep fake detection will lead to advances in these areas.






\end{enumerate}

\bibliography{refs}
\bibliographystyle{model2-names}

\end{document}